\RequirePackage{tikz}
\documentclass[pdflatex,sn-basic]{sn-jnl}
\usepackage{hyperref}

\usepackage{mathtools}
\usepackage{caption}
\usepackage{subcaption}
\usepackage{eurosym}
\usepackage{comment} 
\usepackage{enumitem}
\usepackage{float}
\usepackage{pdfpages,pgfplots,pgfplotstable}
\pgfplotsset{compat=1.8}
\usepgfplotslibrary{statistics}
\usepackage{balance}
\usepackage[title]{appendix}

\DeclarePairedDelimiter\norm{\lVert}{\rVert}  
\DeclarePairedDelimiter\abs{\lvert}{\rvert}  
\DeclarePairedDelimiter\card{\lvert}{\rvert}  

\DeclareMathOperator*{\argmin}{arg\,min} 

\newtheorem{thm}{Theorem}
\newtheorem{lemma}{Lemma}

\newtheorem{prop}[thm]{Proposition}

\begin{document}

\title[Online Decision Making for Trading Wind Energy]{Online Decision Making for Trading Wind Energy}

\author[1]{\fnm{Miguel Angel} \sur{Mu\~noz}}\email{miguelangeljmd@uma.es} 
\author*[2,3]{\fnm{Pierre} \sur{Pinson}}\email{p.pinson@imperial.ac.uk}
\author[4]{\fnm{Jalal} \sur{Kazempour}}\email{jalal@dtu.dk}

\affil*[1]{\orgdiv{OASYS Group}, \orgname{University of Malaga}, \orgaddress{\city{Malaga}, \country{Spain}}}
\affil[2]{\orgdiv{Dyson School of Design Engineering}, \orgname{Imperial College London}, \orgaddress{ \city{London}, \country{United Kingdom}}}
\affil[3]{\orgdiv{Department of Technology, Management and Economics}, \orgname{Technical University of Denmark}, \orgaddress{\city{Kgs. Lyngby}, \country{Denmark}}}
\affil[4]{\orgdiv{Department of Wind and Energy Systems}, \orgname{Technical University of Denmark}, \orgaddress{\city{Kgs. Lyngby}, \country{Denmark}}}

\abstract{We propose and develop a new algorithm for trading wind energy in electricity markets, within an online learning and optimization framework. In particular, we combine a component-wise adaptive variant of the gradient descent algorithm with recent advances in the feature-driven newsvendor model. This results in an online offering approach capable of leveraging data-rich environments, while adapting to the nonstationary characteristics of energy generation and electricity markets, also with a minimal computational burden. The performance of our approach is analyzed based on several numerical experiments, showing both better adaptability to nonstationary uncertain parameters and significant economic gains.

}

\keywords{Decision making under uncertainty, Online learning, Electricity market, Newsvendor model}


\maketitle

\vspace{8mm}
\section{Introduction}
\label{sec:introduction}


\subsection{Problem statement}

Traditionally, the way in which trading wind energy has been considered relied on a two-step approach. These start with the predictive modeling of future energy generation (within either deterministic or probabilistic frameworks). Such forecasts are subsequently used as input to expected utility maximization strategies or, alternatively, some more general forms of optimization problems, e.g., within a stochastic framework and accommodating risk aversion. Although fruitful, these methodologies may be computationally expensive. \textcolor{black}{As a representative recent example, for a scenario-based stochastic optimization setup to offer in electricity markets, \citet{Kraft2023} mentions that computational costs may reach 3 hours for a single trading instance.} In addition, the value of the final decisions is highly affected by the quality of the forecasts employed. \textcolor{black}{This fact was looked at for the general case of newsvendor problems (which are the type of stochastic opimization problems at hand here) by \citet{maggioni}, while a detailed investigation of the impact of forecast quality on optimization in electricity markets (though, not exactly for market participation problems), was detailed in \citet{Ordoudis2016}.} As a consequence, it may be beneficial to integrate the forecasting and decision-making steps, within a so-called \textit{prescriptive analytics} framework \citep{bertsimas2019}. In parallel, electricity markets are amid rapid transformations towards reducing granularity and lead times, facilitating the integration of non-dispatchable energy sources but increasing the computational and adaptability requirements of the offering algorithms.  

In a data-rich and nonstationary environment, approaches relying on online learning and online convex optimization are of direct relevance. For a very complete introduction to these topics, the reader is referred to \citet{shalev2012online}. \textcolor{black}{On the one hand, online learning algorithms free the decision-maker from most assumptions about the wind or market dynamics, since it does not require specific probabilistic forecasts or models about such dynamics. This is more generally the case for a broad range of prescriptive analytics approaches that bypass the forecasting step.} \textcolor{black}{On the other hand, online learning algorithms are typically efficient methods capable of adapting to the increasing computational needs (as will be illustrated by the numerical case study in this paper)}. Furthermore, the online learning analysis is based on regret as opposed to the classical maximization of the expected utility, possibly allowing to derive additional insights into the properties of trading strategies.

\subsection{Status quo with trading wind energy and underlying newsvendor problems}

Most wind energy is traded in wholesale electricity markets (referred to as forward markets in this paper), where an offer is submitted prior to the actual delivery of energy. However, the stochastic nature of wind energy entails incurring deviations from the original offer. There are countless ways of approaching this problem depending on the market structure and how uncertainty is accommodated, and therefore, it is infeasible to fully address such a vast literature. \textcolor{black}{However, let us provide an overview in the following. As a starting point, and since there is no single authoritative review that covers this topic of renewable energy offering in electricity markets, we refer the reader to \citet{Morales2014book}, where the authors study different market variants and strategies assuming a classical stochastic programming framework, as well as \citet{Conejo2010}, which introduces general concepts of decision-making under uncertainty within electricity markets. We deal, in particular, with markets with a dual-price settlement for imbalances, under which there is no possibility of benefiting from a deviation and where imbalance penalties are asymmetric.}

\textcolor{black}{Early works in this area proposed an optimal quantile strategy based on probabilistic forecasts for wind energy production \citep{bremnes2004probabilistic}. Specifically, \citet{pinson2007trading} showed that, in its simplest version of a risk-neutral wind farm without any other assets (e.g., storage, conventional generation), the offering problem necessarily takes the form of a newsvendor problem. Various generalizations were explored by others. \citet{Zugno2013} proposed constraining the offer in both power and probability spaces in order to accommodate risk aversion and behavioral aspects of trading (e.g., anchoring effects towards traditional single-valued forecasts). In parallel, \citet{Mazzi2016} devised and tested a reinforcement learning algorithm to track the optimal quantile in a nonstationary environment. Similarly, \citet{Dent2011} revisited the problem by accounting for the possibility of a population-based price-making behavior. And, for more complex versions of the offering problems, one can revert to a stochastic programming setup \citep{Morales2010}, for instance, owing to inter-temporal constraints, or risk-aversion. If generally considering market offering problems where renewable energy producers are not price-takers (i.e., their decision can then affect market outcomes), \citet{Baringo2013}, as well as \citet{Zugno2013b}, have proposed approaches based on bilevel optimization. Recently, \citet{Kakhbod2021} have investigated the population effect of renewable energy producers and how this affects their offering strategies.} Even though these varied approaches explore alternative angles to generalizing the underlying newsvendor problems in wind energy offering in electricity markets, they still require a two-step procedure (i.e., ``predict, then optimize"). \textcolor{black}{In contrast, a prescriptive approach does not require a forecasting step, since it directly goes from input data to decision. Consequently, there is no need to describe future  wind power generation and market quantities. Hence, no assumption is made about their dynamics.}



Inspired by new advances in decision making under uncertainty in data-rich environments, this problem regained interest in recent years within a prescriptive analytics framework (hence, by integrating forecasting and optimization steps). As a representative example, \citet{stratigakos2021prescriptive} used an ensemble of decision trees that considers the objective function to estimate the energy production. From the modeling perspective, the work of \citet{munoz2020feature} is one of the closest to ours, also aligned with the new stream of research that utilizes features to produce context-specific decisions in a fully data-driven environment. They built upon recent advances with data-driven newsvendor problems \citep{ban2019big}, and proposed an approach that iteratively solves a linear optimization problem to update offering decisions. Although relatively inexpensive, the computation time involved may become an issue in electricity markets like the Australian NEM\footnote{Australian National Electricity Market (NEM). See \url{https://aemo.com.au/}}, where trading and dispatching is based on 5-minute time steps and updates. Moreover, this approach seems redundant in the sense that the complete optimization problem is solved at each and every trading session, even though consecutive training sets may only differ by one or a few samples. Such pitfalls motivates our proposal to explore alternative approaches to wind energy offering in electricity markets.

\subsection{From optimization to online learning}

\textcolor{black}{Instead of using optimization directly, we introduce an offering approach within an online learning paradigm. Online learning can be seen as a special case of online convex optimization (OCO -- considering convex loss functions only) where, instead of tracking optimal decisions, one adaptively and recursively estimate parameters of decision rules (often also referred to as policies). Decision rules are functions that yield decisions based on values of relevant input features. For an introduction to online optimization, we refer the reader to the surveys of \citet{shalev2012online} and \citet{hazan2016introduction}. In addition, for the case of online learning, a recent extensive textbook-like coverage is given by \citet{orabona2019modern}.}

\textcolor{black}{Within OCO, we place emphasis on algorithms that continuously update variables based on gradients (or subgradients) of a convex objective function. Whenever new values of input features and outcomes become available, these algorithms make a step along the gradient, towards the optimum. They ideally accommodate problems for which a closed-form  expression to evaluate the sub-gradient exists (and fast to compute) \citep{duchi2011adaptive, zheng2011gradient}. The well-known online gradient descent approach can be traced back to \cite{zinkevich2003online} and inspired many further developments. Among those are numerous applications within power system operation and electricity markets \citep{gan2016online, hauswirth2017online, colombino2019online, guo2021online,survey}. These methods offer long-term regret guarantees \citep{orabona2019modern}.}

\textcolor{black}{Within the frame of decision-making under uncertainty, the strategy followed by online gradient descent algorithms is in sharp contrast with optimization approaches. These latter approaches solve an independent optimization problem with a different training set (a batch of data) every time a decision has to be updated, e.g., the parameter of the decision rule in \cite{munoz2020feature}. Under convexity assumptions, an optimal solution can be found to each optimization problem, meaning that no single decision can ever achieve better performance on average in that training set. However, there is no certainty that the out-of-sample performance of such a decision enjoys the same privilege in finite sample sets. Instead, only probability guarantees can be offered even if the samples are i.i.d. \citep{van2021data}.}

\textcolor{black}{Indeed, when the underlying data generating processes are nonstationary, the out-of-sample performance can be very poor. This issue can be partly compensated by using a rolling window setting \citep{bashir2018day} that updates the variables frequently. However, there can also be substantial changes within the training set. In that case, the performance of batch optimization approaches may be affected by old samples that do not reflect current conditions. On the contrary, online gradient descent algorithms update the parameters of decision rules through a point-wise update that involves the most recent information only, which enables capturing changes in the characteristics of the underlying data generating processes. Therefore, online gradient methods do not only offer computational advantages. They may also outperform established approaches, e.g., using linear programming with contextual information (even if using a sliding window scheme). This is illustrated based on the toy model examples in Section~\ref{sec:illustrative_examples}, as well as the case study in Section~\ref{sec:case_study}. Their superiority eventually is in terms of both (i) better tracking of the optimal solution within a nonstationary environment, as well as (ii) an increase in market revenues.}

\subsection{Contributions and structure}

The Australian NEM is an example of the existing trend towards shortening lead times and increasing granularity in electricity markets. These developments reduce operational and forecast uncertainty, hence facilitating the integration of stochastic renewable energy sources\footnote{Increasing time granularity in electricity markets, innovation landscape brief, International Renewable Energy Agency (IRENA), Report, 2019}. At the same time, they increase computational needs and require methodologies that adapt to changes in rapid manner. To face these new challenges, we propose an algorithm that combines a feature-driven newsvendor model inspired by \cite{ban2019big} with a variant of the online gradient descent algorithm presented in \cite{zeiler2012adadelta}. We conceive a case study in which we analyze an hourly forward market that closes just before the start of the next period. It relies on actual data from the Danish Transmission System Operator (TSO), Energinet\footnote{See \url{https://energinet.dk/}}, and provide a relevant test bench to illustrate and discussion the salient features of our approach. To the best of our knowledge, this is the first paper that analyzes the problem of trading wind energy in an online learning setting. The contributions of our work are threefold:

\begin{itemize}
    \item we develop an online offering algorithm within an online learning framework. Results show that this algorithm is computationally inexpensive and achieves substantial economic profits;
    \item we propose a new nonstationary regret benchmark against which we empirically compare our algorithm;
    \item we showcase the ability of the proposed algorithm to adapt to nonstationary scenarios through a concise illustrative example. In addition, we analyze the superior economic performance and computational efficiency of our approach based on a case study using real-world data (published by the Danish TSO, Energinet) for a period of more than five.
\end{itemize}

The remaining of the manuscript is structured as follows: Section \ref{sec:nv_rolling_win} introduces the problem of a wind farm offering in the forward market, and for which balancing using a two-price imbalance settlement. Section \ref{sec:online} develops a new offering algorithm based on an adaptive gradient descent algorithm and explores several performance metrics. Section \ref{sec:illustrative_examples} is built upon two illustrative examples that investigate the behavior of an alternative online implementation and the dynamic response of this algorithm in comparison with previous rolling window approaches. Section \ref{sec:case_study} empirically analyzes the performance of our proposed algorithm in a case study based on real data retrieved from the Danish TSO, Energinet. Finally, conclusions and perspectives for future work are gathered in Section \ref{sec:conclusions}.

\section{Preliminaries}
\label{sec:nv_rolling_win}

\subsection{Mathematical notations}

\textcolor{black}{We introduce here some of the most relevant mathematical notations used throughout the paper. These are placed into context when further describing the optimization and learning problems at hand in the following. In terms of indices and sets, we use $j$ as an index for features and auxiliary information, while $t$ is an index for time periods (hours in practice, or programme time units in the electricity market of interest). These time indices are gathered within 2 sets $\mathcal{T}^{\rm{in}}$ and  $\mathcal{T}^{\rm{oos}}$, which are for training (in-sample) and testing (out-of-sample), respectively.}

\textcolor{black}{When looking at newsvendor problems and offering in electricity markets, key parameters include $\psi^{+}_t$, the marginal opportunity cost for overproduction at hour $t$ (\euro/MWh), and $\psi^{-}_t$, the marginal opportunity cost for underproduction at hour $t$ (\euro/MWh). These are defined based on $\lambda^{\rm{F}}, \lambda^{\text{UP}}$, and $\lambda^{\text{DW}} \in \mathbb{R}$, which are the forward, up-regulation and down-regulation prices, respectively. In terms of the renewable energy producer, the asset or portfolio at hand has a nominal capacity $\overline{E}$, also translating to a maximum offer in terms of energy in the market for each and every programme time unit (hence, we express $\overline{E}$ in MWh eventually). The decision variable is then the energy bid $E^{\rm{F}}_t$ (MWh) for that time, while the amount of energy actually produced is $E_t$ (MWh). Within our data-driven framework, that decision is based on a vector $\mathbf{x}_t$ of auxiliary information (i.e., features), associated to a decision rule vector $\mathbf{q}_t$.}

\textcolor{black}{Finally, for the type of online learning approach described in the following, the method and resulting algorithm rely on the gradient or subgradient of the objective function at hand, which we denote by $\mathbf{g}_t$, as well as a dynamic learning vector $\boldsymbol{\eta}_t$. We write $g_{t,j}$ and $\eta_{t,j}$ the $j^{\text{th}}$ components of these vectors. The algorithm has a number of hyperparameters involved, i.e., $\mu$ as a forgetting factor to temporally smooth the marginal opportunity costs $\psi^{+}_t$ and $\psi^{-}_t$, $\eta$ to control the learning rate, $\alpha$ to smooth the discontinuity in the derivative of the pinball loss function, and $\rho$ as a decay constant  that controls the adaptation to new gradients. A strictly positive, though small, constant $\epsilon$ is used in the definition of the dynamic learning vector $\boldsymbol{\eta}_t$ in order to avoid dividing by 0.}

\subsection{Newsvendor problem on a rolling time-window} 
We first introduce the problem of a wind farm offering in a forward market, which is cleared some time before their actual production is realized. Therefore, the producer is likely to suffer deviations from her offer. These are settled \emph{ex-post} in a real-time (balancing) market under a two-price imbalance settlement mechanism. Furthermore, the offer is assumed to be always accepted, as the marginal operational cost of wind farms is close to zero and therefore this technology is usually prioritized for being scheduled. The eventual market revenue $\rho \in \mathbb{R}$ of a wind farm is given by the summation of the amounts obtained in the forward ($\rho^{\rm{F}}$) and in the balancing markets ($\rho^{\rm{B}}$), i.e.,
%
%
\begin{equation}\label{eq:revenue_lamb}
        \rho = \rho^{\rm{F}} + \rho^{\rm{B}} =  \lambda^{\rm{F}} E^{\rm{F}}  - \lambda^{\text{UP}} (E^{\rm{F}} - E)^{+} + \lambda^{\text{DW}}  (E - E^{\rm{F}})^{+}  \, ,
\end{equation}
where $(a)^{+}= \max(a,0)$. In addition, the unknown parameters $\lambda^{\rm{F}}, \lambda^{\text{UP}}$, and $\lambda^{\text{DW}} \in \mathbb{R}$ are the forward, up-regulation and down-regulation prices, respectively. The key decision variable for the wind farm is her offer $E^{\rm{F}} \in \mathbb{R}^+$ at the forward stage. Note that $E \in \mathbb{R}^+$ denotes the actual realization of her stochastic energy production, which is obviously unknown at the forward stage. In accordance to \eqref{eq:revenue_lamb}, the revenue ($\lambda^{\rm{F}} E^{\rm{F}}$) from the forward stage is then altered when the producer deviates from her offer $E^{\rm{F}}$. When the production is greater than expected $E \ge E^{\rm{F}}$, the producer is to sell excess energy generation $E - E^{\rm{F}} > 0$ at the downward regulation at price $\lambda^{\text{DW}}$. On the contrary, if she produces less than her forward offer $E \le E^{\rm{F}}$, the wind farm has to buy the missing energy $E^{\rm{F}} - E > 0$ at teh upward regulation at price $\lambda^{\text{UP}}$.
Under a two-price imbalance settlement, one has $\lambda^{\text{UP}} \ge \lambda^{\rm{F}}$ and $\lambda^{\text{DW}} \le \lambda^{\rm{F}}$, with at most one of them different from $\lambda^{\rm{F}}$ \cite[Ch. 7]{Morales2014book}.
In accordance with the aforementioned description, let  $\psi^{+}, \psi^{-} \in \mathbb{R}^+$ denote penalties for over- or under-production as
%
\begin{align} 
    \psi^{+} &= \lambda^{\rm{F}} - \lambda^{\text{DW}}, \label{eq:psi_pls}\\
    \psi^{-} &= \lambda^{\text{UP}} - \lambda^{\rm{F}}  \, . \label{eq:psi_min}
\end{align}
Using \eqref{eq:psi_pls} and \eqref{eq:psi_min} and the equivalence $E - E^{\rm{F}} = (E - E^{\rm{F}})^{+} - (E^{\rm{F}} - E)^{+}$, we  reformulate \eqref{eq:revenue_lamb} as 
\begin{equation}\label{eq:revenue_psi}
        \rho = \lambda^{\rm{F}} E - \biggl( \psi^{+} (E - E^{\rm{F}})^{+} + \psi^{-} (E^{\rm{F}} - E)^{+} \biggr)  \, . 
\end{equation}

%


Note that the first term of \eqref{eq:revenue_psi} is out of the control of the price-taker wind farm, as both $\lambda^{\rm{F}}$ and $E$ are uncertain parameters. Therefore, the profit-maximizing offer $E^{\rm{F}^{*}}$ of the wind farm in the forward market can be computed by minimizing the expected deviation cost as
%
\begin{align}\label{eq:nv_stoch}
    E^{\rm{F}^{*}} = \argmin_{E^{\rm{F}} \in [0, \overline{E}]} \enskip  \mathbb{E} \biggl[  \psi^{+}(E - E^{\rm{F}})^{+} + \psi^{-}(E^{\rm{F}} - E)^{+} \biggr],
\end{align}
where $\mathbb{E}[\cdot]$ is the expectation operator. The optimization program \eqref{eq:nv_stoch} solves an instance of the very well-studied newsvendor model \citep{newsvendor1}. Under a price-taker scenario, i.e., when the market participant's decision are assumed not to affect market outcomes, an analytical solution to \eqref{eq:nv_stoch} can be computed with \citep{bremnes2004probabilistic, pinson2007trading}
\begin{equation}\label{eq:nv_analitical_sol}
    E^{\rm{F}^{*}} = F^{-1}_{E}\left(\frac{\bar{\psi}^{+}}{\bar{\psi}^{+}+\bar{\psi}^{-}}\right),
\end{equation}
%
%
where $F^{-1}_{E}(.)$ is the cumulative distribution function (cdf) of the renewable energy production and the overline denotes the expected value of the random variable (estimated as the average over available data). The reader is referred to \cite{maggioni} for a discussion about the value of right distribution in newsvendor applications. 

On top of the fact that the true distribution of the wind production and the optimal quotient are generally unknown, \eqref{eq:nv_analitical_sol} suffers from another major drawback, which is its inability to directly profit of additional information that may be available, e.g., wind energy forecasts for neighboring areas, or additional information about the state of the electricity market. In fact, it is usually the case that the wind farm operator has access to a vector of \textit{auxiliary information}, also known as \textit{features} $\mathbf{x} \subseteq \mathcal{X} \in \mathbb{R}^p$, where $p$ denotes the dimension of the feature vector. This feature vector may help explaining the behavior of the uncertain parameters in \eqref{eq:nv_stoch}. As proposed by \cite{ban2019big}, this information can be exploited in newsvendor instances assuming that the optimal offer follows a linear decision rule of the form $ E^{\rm{F}}: \mathcal{X} \rightarrow \mathbb{R}$, $E^{\rm{F}} = \mathbf{x}^\top \mathbf{q}$ with $\mathbf{q} \in \mathbb{R}^p$ being a decision vector that parameterizes the linear model. This decision rule can easily reproduce an intercept setting a component of the feature vector $\mathbf{x}$ equal to one. Then, considering that a set of historical samples $\left\{(E_t, \psi_t^{-}, \psi_t^{+},  \mathbf{x}_t), \forall t\in\mathcal{T}^{\rm{in}} \right\}$ is available, we compute the best decision $\mathbf{q}^{\text{LP}}$ for this set by solving the following linear program:
%
%
%
%
%
%
\begin{subequations} \label{eq:nv_rolling_win_opt} 
\begin{align}
    \mathbf{q}^{{\text{LP}}^{*}} = \argmin_{\mathbf{q}} & \enskip \frac{1}{\card{\mathcal{T}^{\rm{in}}}} \sum_{t\in\mathcal{T}^{\rm{in}}} \psi_{t}^{+}\left(E_t - \mathbf{x}^\top_t \mathbf{q} \right)^{+} + \psi_{t}^{-}\left(\mathbf{x}^\top_t \mathbf{q} - E_t\right)^{+} \label{eq:nv_dis_feature1}\\
     \text{s.t.} & \enskip 0 \leq \mathbf{x}^\top_t \mathbf{q} \leq \overline{E}, \enskip \forall t\in\mathcal{T}^{\rm{in}}, \label{eq:nv_dis_feature2}
\end{align}  
\end{subequations}
where $\card{\cdot}$ denotes the cardinality of a set. Note that this model does not implicitly assume a price-taker scenario. In fact, correlations between penalties and wind features may be captured in systems with high wind power penetration. Although the linear structure of the mapping may seem restrictive, more complex relationships can be obtained by transforming the feature space, e.g., using a Taylor approximation \citep{ban2019big} or a spline basis. Next, by defining the box projection  
\begin{align}
    \pi(\mathbf{x}, \mathbf{q}) = \min \bigg(\max(0, \mathbf{x}^\top \mathbf{q}), \overline{E}\bigg), \label{eq:nv_rolling_win_eval}
\end{align}
the optimal offer derived from new contextual information $\mathbf{x}_{t'}$ can be computed as $E^{\rm{F}}_{t'} = \pi(\mathbf{x}_{t'}, \mathbf{q}^{\text{LP}})$. As discussed in \cite{munoz2020feature}, when new points are incorporated into the dataset $\mathcal{T}^{\rm{in}}$, the problem \eqref{eq:nv_rolling_win_opt} can be iteratively solved to update the value of $\mathbf{q}^{\text{LP}}$. In the remaining of the manuscript, we refer to this approach as LP (from Linear Programming).

\section{Online learning in newsvendor problems}\label{sec:online}


In the Online Convex Optimization (OCO) framework, a decision-maker faces an online learning problem where iterative decisions are to be made. The cost of each decision is determined by a convex loss function $f_t: \mathbb{R}^{d_z} \rightarrow \mathbb{R}$ unknown beforehand. After a decision $\mathbf{z}_t \in Z \subseteq \mathbb{R}^{d_z}$ is made, the decision-maker learns $f_t$ and pays $f_t(\mathbf{z}_t)$. Within OCO the Online Gradient Descent (OGD) algorithm, introduced by \cite{zinkevich2003online}, has proven to be very effective and versatile \citep{gan2016online, narayanaswamy2012online,  hauswirth2016projected, nonhoff2020online, Wood2021online}. Starting from an initial value, the OGD performs iterative updates $\mathbf{z}_t$ based on (sub-)gradients of $f_t$, denoted as $\mathbf{g}_t$ from hereon. The magnitude of the step is controlled through a variable learning rate $\eta_t$. On each round, the updated vector is forced to lie within the feasible region $Z$ through the Euclidean projection. In the OGD we rely on just the last point learned to obtain a gradient, thus resulting in a computationally inexpensive method, especially if the gradient and projection can be computed through a closed-form expression.

The selection of the learning rate is of paramount importance. The original proposal by \cite{zinkevich2003online} presents two main alternatives, namely, a variable and a fixed learning rate. In a dynamic environment, the classical choice $\eta_t \in \mathbb{R}^+$, $\eta_t \propto t^{-1/2}$ where $\propto$ denotes the proportional operator, is not suitable due to the fact that $\lim_{t \rightarrow \infty} = 0$, reducing the ability to track changes as $t$ increases. Alternatively, one could select a fixed value $\eta_t = \eta$ that keeps this capacity unaltered but may lose the fast convergence that the initial high values of $\eta_t$ provide. Regardless of the selection, both choices are scale-dependent and treat each component of the gradient vector equally. To tackle this, \cite{mcmahan2010adaptive} and \cite{duchi2011adaptive} propose to use a component-wise adaptive rate $\boldsymbol{\eta}_t \in \mathbb{R}^p$ and $\eta_{t,j} = \eta (\sum_{k=1}^t g_{k,j}^2)^{-1/2}$ where $g_{t,j}$ is a component of the gradient vector $\mathbf{g}_t = [g_{t,1}, ..., g_{t,j}, ..., g_{t,p}]^{\top}$. As in the case of $\eta_t \propto t^{-1/2}$, the previous expression is monotonically decreasing (component-wise), again limiting the long-term ability to learn. Aware of this limitation, \cite{zeiler2012adadelta} suggests an exponentially decaying average of the squared gradients to modulate the learning rate based on the most recent information. We employ this gradient descent variant to implement our algorithm in Section \ref{subsec:ogd_implementation}.

In the online learning community, the \emph{de facto} metric to evaluate the performance of a series of decision vectors $\mathbf{z}_1, ..., \mathbf{z}_T$ is the regret $\mathcal{R}_T \in \mathbb{R}$. The regret provides a versatile and, in a sense, normalized metric to compare an algorithm through different problems with the advantage that little assumption is made about the oracle that generates the decisions. Traditionally, the benchmark used to compute regret is the best single action in hindsight that can be obtained as the solution to an offline optimization problem under perfect information. However, in a dynamic environment, this benchmark can be beaten easily. In Section \ref{subsec:performance_eval} we propose an alternative benchmark more suitable for the nonstationary context of the wind energy problem.




\subsection{Online newsvendor} \label{subsec:ogd_implementation}

In this section, we particularize the gradient descent introduced in the previous paragraphs to the context of the wind farm offering in a forward market, incorporating elements of the rolling window problem presented in Section \ref{sec:nv_rolling_win}. We name the resulting algorithm OLNV (from OnLine NewsVendor). Contrary to the rolling window approach, the OLNV algorithm updates $\mathbf{q}$ based on the information provided by the last realization. The objective function  \eqref{eq:nv_dis_feature1} when the set $\mathcal{T}^{\rm{in}}$ reduces to one sample yields
\begin{align}
    NV_t(\mathbf{q}) =  \psi_{t}^{+}\left(E_t - \mathbf{x}^\top_t \mathbf{q} \right)^{+} + \psi_{t}^{-}\left(\mathbf{x}^\top_t \mathbf{q} - E_t\right)^{+}.\label{eq:nv_ob_online}
\end{align}

The OLNV method requires computing a gradient of the objective function, for which we analyze two alternative procedures in the following paragraphs. 

The first approach is inspired by the work of \cite{zheng2011gradient} on the pinball loss, a particular case of the objective function found in newsvendor models. Since the pinball loss is not strictly differentiable, the authors propose an alternative smooth approximation to ensure that computing gradients is always possible. Note that in our case the objective function~\eqref{eq:nv_ob_online} is not differentiable at $E_t = \mathbf{x}_t^\top \mathbf{q}$. Therefore, we first propose to circumvent this issue extending the approach in \cite{zheng2011gradient} to the more general expression~\eqref{eq:nv_ob_online} that considers arbitrary (positive) penalties as 
%
\begin{align}
    NV_{t, \alpha}(\mathbf{q}) &= \psi^{+}_t (E_t - \mathbf{x}^\top_t \mathbf{q}) + \alpha (\psi^{+}_t + \psi^{-}_t) \log (1 + e^{ - (E_t - \mathbf{x}^\top_t \mathbf{q}) / \alpha})  \, , \label{eq:nv_smooth_obj}
\end{align}
where $\alpha > 0$ is a parameter that controls the approximation and where higher values of this parameter result in smoother functions. The function $NV_{t, \alpha}$ is convex in $\mathbf{q}$ and upper bounds $NV_t$ for any value of $\mathbf{q}$ as proven in Propositions~\ref{prop:smooth_nv_convexity} and \ref{prop:nv_approximation} in Appendix~\ref{sec:appendixA}, respectively. Then, we derive a closed-form solution to obtain gradients of \eqref{eq:nv_smooth_obj}, yielding
\begin{align} \label{eq:nv_gradient}
    \nabla NV_{t, \alpha}(\mathbf{q}) &= \bigg(-\psi^{+}_t + (\psi^{+}_t + \psi^{-}_t) \frac{1}{1 + e^{(E_t - \mathbf{x}^\top_t \mathbf{q}) / \alpha}} \bigg) \mathbf{x}_t  \, .
\end{align}

The second approach deals directly with the objective function as formulated in \eqref{eq:nv_ob_online}. Even though the original objective is not strictly differentiable, a variant of the OLNV algorithm is readily applicable to subdifferentiable functions, provided that a subgradient can be computed instead \citep{orabona2019modern}. In this case, the mapping that returns a subdifferential of \eqref{eq:nv_ob_online} is given by
\begin{align} \label{eq:nv_subgradient}
    \partial NV_t(\mathbf{q}) = \begin{cases}
        - \psi^{+}_t \mathbf{x}_t, &  E_t - \mathbf{x}^\top_t \mathbf{q} > 0, \\
        \psi^{-}_t \mathbf{x}_t, &  E_t - \mathbf{x}^\top_t \mathbf{q} < 0, \\
        [-\psi^{+}_t \mathbf{x}_t, \psi^{-}_t \mathbf{x}_t], &  E_t - \mathbf{x}^\top_t \mathbf{q} = 0  \, . \\
    \end{cases}
\end{align}
Note that, when $E_t - \mathbf{x}^\top_t \mathbf{q} = 0$, any value in the interval $[- \psi^{+}_t \mathbf{x}_t, \psi^{-}_t \mathbf{x}_t]$ is a legitimate subgradient belonging to $\partial NV_t(\mathbf{q})$. For the sake of simplicity and reproducibility, the implementation of our algorithm returns zero whenever this condition is fulfilled. 

Once a gradient as in~\eqref{eq:nv_gradient} or a subgradient as in \eqref{eq:nv_subgradient} has been computed, the key step of OLNV is to update $\mathbf{q}_t$ using a multidimensional learning rate $\boldsymbol{\eta}_t \in \mathbb{R}^p$ through
\begin{align}
    \mathbf{q}_{t+1} &= \Pi(\mathbf{q}_{t} - \boldsymbol{\eta}_t \circ \mathbf{g}_t, \mathbf{x}_t)  \, , \label{eq:subgr_update}
\end{align}
where $\circ$ denotes the element-wise product, $\mathbf{g}_t = \nabla NV_{t, \alpha}(\mathbf{q}_t)$ or $\mathbf{g}_t = \partial NV_t(\mathbf{q}_t)$ depending on the implementation of OLNV, and $\Pi$ is a projection operator defined as $\Pi: \mathbb{R}^p \times \mathcal{X} \rightarrow \mathbb{R}^p $. Precisely, $\Pi$ maps its arguments into the solution of the following optimization problem:
\begin{align}
    \tag{$\mathcal{P}$}
    \Pi(\mathbf{o}, \mathbf{x}) = & \argmin_{\mathbf{q} \in Q(\mathbf{x})} \frac{1}{2} \norm{\mathbf{q} - \mathbf{o}}_2  \, ,  \label{eq:nv_projection}  
\end{align}
where $o$ represents a candidate to update the decision vector and is computed $o = \mathbf{q}_{t} - \boldsymbol{\eta}_t \circ \mathbf{g}_t$. The feasible set in \eqref{eq:nv_projection} is defined by the set-valued mapping $Q: \mathcal{X} \rightrightarrows \mathbb{R}^p$, $Q(\mathbf{x}) =  \{\mathbf{q} : 0 \leq \mathbf{x}^\top \mathbf{q}  \leq \overline{E}\}$. Note that, for any input $\mathbf{x}$ the output of $Q$ is a convex region bounded by two parallel hyperplanes. As the Euclidean norm is used, a unique solution is guaranteed to exist for any instance of \eqref{eq:nv_projection}. Generally, the Euclidean projection of a point into a convex set requires solving a convex optimization problem, however the definition of $Q$ allows us to find a closed-form expression, yielding
\begin{align}
    \Pi(\mathbf{o}, \mathbf{x}) = \begin{cases}
        \mathbf{o}, & 0 \le \mathbf{x}^\top \mathbf{o} \le \overline{E}  \, , \\
        \mathbf{o} + \frac{\overline{E} - \mathbf{x}^\top \mathbf{o}}{\norm{\mathbf{x}}^2_2} \mathbf{x}, & \mathbf{x}^\top \mathbf{o} >\overline{E}  \, , \\
        \mathbf{o} + \frac{ - \mathbf{x}^\top \mathbf{o}}{\norm{\mathbf{x}}^2_2} \mathbf{x}, & \mathbf{x}^\top \mathbf{o} < 0  \, .
    \end{cases}
\end{align}
This reduces the resolution of the optimization problem \eqref{eq:nv_projection} to evaluating the above expression. Even though the operator $\Pi$ guarantees the feasibility of $\mathbf{q}_{t}$ under the realization $\mathbf{x}_{t-1}$, we need to resort to \eqref{eq:nv_rolling_win_eval} setting $E^{\rm{F}}_t = \pi(\mathbf{x}_{t}, \mathbf{q}_t)$ to guarantee $E^{\rm{F}}_t$ remains feasible for any new arbitrary $\mathbf{x}_t$. 

%


The last remaining aspect is to compute the vector $\boldsymbol{\eta}_t$ following the ideas in \cite{zeiler2012adadelta}. Let $\mathbf{g}_t = [g_{t,1}, ..., g_{t,j}, ..., g_{t,p}]^{\top}$ be a gradient or subgradient vector computed through \eqref{eq:nv_gradient} and \eqref{eq:nv_subgradient}. Then, we can define the squared running average of each component as
%
\begin{align}
    \overline{g}_{t,j}^2 = \rho \overline{g}_{t-1,j}^2 + (1 - \rho) g_{t,j}^2  \, , \label{eq:grad_exp_decay_ave}
\end{align}
where $\rho \in [0, 1)$ is a decay constant and $\overline{g}_{0,j}^2 = 0$. The auxiliary variable $\overline{g}_{t,j}^2$ is then used to compute the independent learning rate applied to the associated decision vector component following
\begin{align}
    \eta_{t,j} = \frac{\eta}{\sqrt{\overline{g}_{t,j}^2 + \epsilon}}  \, , \label{eq:eta_update_olnv}
\end{align}
where $\epsilon \in \mathbb{R}^+$ helps better conditioning the denominator (by avoiding division by 0)and $\eta > 0$ is a constant. We use the update given by \eqref{eq:grad_exp_decay_ave} and \eqref{eq:eta_update_olnv} in the proposed OLNV algorithm with the values $\epsilon = 10^{-6}$ and $\rho=0.95$, as originally suggested in \cite{zeiler2012adadelta}. The benefits of this update is twofold. On the one hand, OLNV adapts each learning rate component to the scale of the incumbent feature. On the other hand, OLNV tracks the most recent dynamic between the uncertain vector $[E_t, \psi_t^+, \psi_t^-$] and the feature vector $\mathbf{x}_t$. The OLNV algorithm for the feature-driven wind energy trading problem is compiled in Algorithm \ref{alg:nv}.
\begin{algorithm}
\caption{Online Newsvendor (OLNV)}\label{alg:nv}
    \begin{algorithmic}[1]
    \Require Initial values $\mathbf{q}_1 \in \mathbb{R}^p$, $\eta > 0$, $\rho \in [0, 1)$, $\epsilon \in \mathbb{R}^{+}$
    \State Initialize $\overline{g}_{0,j}^2 = 0, \enskip \forall j$ 
    \For{$t = 1$ to $T$}
        \State Output $\mathbf{q}_{t}$
        \State Receive $\mathbf{x}_t$
        \State Compute $E^{\rm{F}}_t = \pi(\mathbf{x}_{t}, \mathbf{q}_t)$
        \State Receive $NV_t$ and pay $NV_t(E^{\rm{F}}_t)$
        \State Set $\mathbf{g}_t = \nabla NV_{t, \alpha}(\mathbf{q}_{t})$ or $\mathbf{g}_t = \partial NV_t(\mathbf{q}_{t})$
        \State Accumulate $\overline{g}_{t,j}^2 = \rho \overline{g}_{t-1,j}^2 + (1 - \rho) g_{t,j}^2, \enskip \forall j$
        \State Compute $\eta_{t,j} = \eta (\overline{g}_{t,j}^2 + \epsilon)^{-1/2}, \enskip \forall j$
        \State Update $\mathbf{q}_{t+1}$ = $\Pi (\mathbf{q}_{t} - \boldsymbol{\eta}_t \circ \mathbf{g}_t, \mathbf{x}_t)$ solving \eqref{eq:nv_projection}
    \EndFor
    \end{algorithmic}
\end{algorithm}

\textcolor{black}{Despite the fact we have considered a single wind farm in the derivation, the proposed OLNV algorithm is general enough to be exploited for an aggregation of wind farms, or in general, for a portfolio of diverse renewable energy sources with uncertain production, just by combining the capacity and generation of the assets. Equally, the potential spatial correlation among production of wind farms does not affect the feasible region of the newsvendor model, and therefore does not complicate the OLNV algorithm.}
\textcolor{black}{On the contrary, adding storage to the generation portfolio forces the model to include inter-temporal constraints that dramatically reshape the feasible region, implying that the current decision will affect future outcomes. In this case, the decision-maker can resort to classical dynamic programming \citep{hargreaves2012commitment} or more advanced learning algorithms such as budget-constrained online learning \citep{liakopoulos2019cautious, sherman2021lazy} or reinforcement learning algorithms \citep{kuznetsova2013reinforcement, sutton2018reinforcement}.}

{\color{black} Finally, even if a population effect may be present for renewables in electricity markets (i.e., even if price-taker individually, the sum of individual actions of these producers may impact market outcomes), several wind power producers can effectively use the OLNV algorithm to improve the profitability of their offer within the same region. The fact that each competing producer has different contextual information available and may process it in alternative ways mitigates possible increases in the volatility of their outcomes that could arise from correlated generation.}

\subsection{Regularization through average penalty anchoring}

In an electricity market with a two-price imbalance settlement scheme, it is common that $\psi^{+}_t=\psi^{-}_t = 0$ over a significant number of hours, meaning that load and generation are close to being balanced. In this situation, from~\eqref{eq:nv_ob_online}, the producer experiences no cost no matter the deviation from the actual production. Moreover, the gradients computed through~\eqref{eq:nv_ob_online} are zero and therefore the variable vector $\mathbf{q}_t$ is not updated, wasting information about the relationship between $E^{\rm{F}}_t$ and $\mathbf{x}_t$. And, when penalties are different from zero, they typically exhibit random behavior with sharp spikes representing highly imbalanced scenarios which, in turn, yields destabilizing updates of the vector $\mathbf{q}_t$. To tackle both issues, we propose performing the following convex transformation of the original penalties:
%
%
\begin{align}
    \psi^{+'}_t &= \mu \psi^{+}_t + (1 - \mu) \overline{\psi}^{+}  \, , \label{eq:co_penalty_up}\\
    \psi^{-'}_t &= \mu \psi^{-}_t + (1 - \mu) \overline{\psi}^{-}  \, , \label{eq:co_penalty_dw}
\end{align} 
where $0 \le \mu \le 1$ and $\overline{\psi}^{+}, \overline{\psi}^{-} \in \mathbb{R}^+$ are the historical average penalties. This convex transformation is inspired by the concept of constraining the optimal offer around the point forecast proposed by \cite{Zugno2013}. In contrast though, we do not impose hard constraints on the decision vector $\mathbf{q}_t$. Instead, we smooth the objective function using as anchor the sample average optimal market quantile determined by the average market penalties $\overline{\psi}^{+}$ and $\overline{\psi}^{-}$. To do so, we consider a convex combination of the original objective function \eqref{eq:nv_dis_feature1} with an additional term that minimizes such a quantile,
%
\begin{align}
    NV_t^{\rm{R}} = &\mu\psi_{t}^{+}\left(E_t - \mathbf{x}^\top_t \mathbf{q} \right)^{+} +\mu \psi_{t}^{-}\left(\mathbf{x}^\top_t \mathbf{q} - E_t\right)^{+}  \nonumber \\
    &+  (1 - \mu) \overline{\psi}^{+}\left(E_t - \mathbf{x}^\top_t \mathbf{q} \right)^{+} + (1 - \mu) \overline{\psi}^{-}\left(\mathbf{x}^\top_t \mathbf{q} - E_t\right)^{+}  \, .
\end{align}

Then, using \eqref{eq:co_penalty_up} and \eqref{eq:co_penalty_dw}, the original objective structure is recovered, i.e.,
\begin{align}
     NV_t^{\rm{R}} &=   \psi^{+'}_t \left(E_t - \mathbf{x}^\top_t \mathbf{q} \right)^{+} + \psi^{-'}_t\left(\mathbf{x}^\top_t \mathbf{q} - E_t\right)^{+}  \, .
\end{align}
Therefore, by replacing $\psi^{+}_t, \psi^{-}_t$ with $\psi'^{+}_t, \psi'^{-}_t$ in the original objective function, we regularize the learning procedure at no extra computational cost. When the available samples are not sufficient to provide reliable estimates of the true $\overline{\psi}^{+}$ and  $\overline{\psi}^{-}$, the producer can resort to assume a balanced market with penalties $\overline{\psi}^{+} = \overline{\psi}^{-} = 1$. Thus, with $\mu < 1$, provided that $\overline{\psi}^{+}, \overline{\psi}^{-} > 0$, the algorithm utilizes the information contained in samples with both penalties equal to zero, potentially accelerating the convergence and obtaining smoother updates through the gradient. The same reasoning applies to the smooth objective function.

\subsection{Performance evaluation} \label{subsec:performance_eval}



In order to assess the economic performance of our algorithm over a set of testing samples $\left\{(E_t, \psi_t^{-}, \psi_t^{+},  \mathbf{x}_t), \forall t\in\mathcal{T}^{\rm{oos}} \right\}$, we use the average deviation cost. To lighten the notation, we write $T = \card{\mathcal{T}^{\rm{oos}}}$. Consider that we have obtained successive offers $E_1^{\rm{F}},..., E_{T}^{\rm{F}}$ over the test set, by using~\eqref{eq:nv_rolling_win_opt} and \eqref{eq:nv_rolling_win_eval} or from Algorithm~\ref{alg:nv}, after iteratively going through all the samples belonging to the test set $\mathcal{T}^{\rm{oos}}$. We then calculate the average deviation cost as 
\begin{align} \label{eq:nv_metric_abs}
    NV^{\rm{oos}} = \frac{1}{T} \sum_{t \in \mathcal{T}^{\rm{oos}}} \psi^{-}_{t} (E_{t} - E^{\rm{F}}_{t})^+{+} \psi^{+}_{t} (E^{\rm{F}}_{t} - E_{t})^+  \, .
\end{align}
The value of this metric gives limited information about how a particular method is performing. A natural benchmark is the score obtained when a forecast of the wind energy production (in the sense of minimizing the root mean square error) is directly used as an offer in the market. We refer to this method as $\text{FO}$ (from FOrecast). Let $NV^{\rm{oos}}_{\text{FO}}$ be the deviation cost incurred by $\text{FO}$. We then redefine the original metric in relative terms, i.e.,
%
\begin{align} \label{eq:nv_metric_rel}
    NV^{\rm{oos}} (\%) = 100 \, \frac{NV^{\rm{oos}}_{FO} - NV^{\rm{oos}}}{NV^{\rm{oos}}_{FO}} \, .
\end{align}
Consequently, the metric expresses an improvement (as a percentage), where a value of 100\% means perfect performance with zero deviation cost. 

For online learning problems the customary performance measure is the regret. Traditionally, the regret compares a sequence of decision $\mathbf{q}_1,..., \mathbf{q}_{T}$ against the best single vector in hindsight $\mathbf{q}^{\mathcal{H}}$. The latter is computed \emph{ex-post} solving a problem analogous to \eqref{eq:nv_rolling_win_opt} once the whole collection of samples belonging to $\mathcal{T}^{\rm{oos}}$ is known. Let $Q^{\mathcal{H}}$ be the intersection of all feasible sets $Q(\mathbf{x}_t)$, more precisely  $Q^{\mathcal{H}}: \mathcal{X} \rightrightarrows \mathbb{R}^p$, $Q^{\mathcal{H}} = \{\mathbf{q} : 0 \leq \mathbf{x}_t^\top \mathbf{q}  \leq \overline{E}, t \in \mathcal{T}^{\rm{oos}}\}$. The \emph{static} regret is
\begin{align}
    \mathcal{R}_T^s &= \sum_{t \in \mathcal{T}^{\rm{oos}}} NV_t(\mathbf{q}_t) - \min_{\mathbf{q} \in Q^{\mathcal{H}}} \sum_{t \in \mathcal{T}^{\rm{oos}}} NV_t(\mathbf{q})  \, . \label{eq:regret_nv_stdr}
\end{align}

Given the assumption of a nonstationary environment, outperforming a constant $\mathbf{q}^{\mathcal{H}}$ can be a relatively easy task even though it is determined under perfect information. Alternatively, one may consider the \emph{worst-case} regret \citep{besbes2015non} interchanging the sum and minimum, i.e.,
\begin{align}
    \mathcal{R}_T^w &= \sum_{t \in \mathcal{T}^{\rm{oos}}} NV_t(\mathbf{q}_t) - \sum_{t \in \mathcal{T}^{\rm{oos}}} \min_{\mathbf{q} \in Q(\mathbf{x}_t)} NV_t(\mathbf{q})  \, , \label{eq:regret_nv_worst}
\end{align}
where the second term of \eqref{eq:regret_nv_worst} gives the best individual decision $\mathbf{q}_t^{\mathcal{H}} \in \argmin_{\mathbf{q} \in Q(\mathbf{x}_t)} NV_t(\mathbf{q})$. The regret computed in this way can be very pessimistic and unrealistic. Note that in the context of the wind farm, it is always possible to find a value for $\mathbf{q}$ such that $E_t - \mathbf{x}_t^{\top} \mathbf{q} = 0$, and therefore~\eqref{eq:regret_nv_worst} readily reduces to the summation of the original objective function $\mathcal{R}^w_T = \sum_{t \in \mathcal{T}^{\rm{oos}}} NV_t(\mathbf{q}_t)$. Alternatively, \cite{zinkevich2003online} proposed to compare the performance of online algorithms against a sequence of arbitrary decisions $\mathbf{u}_1, ..., \mathbf{u}_T$, $\mathbf{u}_t \in Q(\mathbf{x}_t)$,
\begin{align}
    \mathcal{R}_T^d &= \sum_{t \in \mathcal{T}^{\rm{oos}}} NV_t(\mathbf{q}_t) - \sum_{t \in \mathcal{T}^{\rm{oos}}} NV_t(\mathbf{u}_t)  \, . \label{eq:regret_nv_sequence}
\end{align}

We refer to this approach as \emph{dynamic} regret. This formulation allows to define a metric with an adjustable difficulty between the previous benchmarks. Note that \eqref{eq:regret_nv_stdr} and \eqref{eq:regret_nv_worst} are special cases of \eqref{eq:regret_nv_sequence} with $\mathbf{u}_t = \mathbf{q}^{\mathcal{H}} \enskip \forall t$ and $\mathbf{u}_t = \mathbf{q}_t^{\mathcal{H}}\enskip \forall t$, respectively. Then, the question is how to choose a reasonable series of reference benchmarks $\mathbf{u}_t$ to use against OLNV. To this end, we propose dividing $\mathcal{T}^{\rm{oos}}$ in $k$ adjacent partitions of equal length $l$, except possibly the last one. Without loss of generality, by assuming  $T - k l = 0$, we have $\mathcal{T}^{\rm{oos}}_i = \{t: (i-1) l + 1 \le t \le i l\}, i = 1, ..., k$. Let us define the feasible sets $Q^{\mathcal{H}}_i = \{\mathbf{q} : 0 \leq \mathbf{x}_t^\top \mathbf{q} \leq \overline{E}, t \in \mathcal{T}^{\rm{oos}}_i\}$. Accordingly, we can compute $\mathbf{q}^{\mathcal{H}}_i = \argmin_{\mathbf{q} \in Q^{\mathcal{H}}_i} \sum_{t \in \mathcal{T}^{\rm{oos}}_i} NV_t(\mathbf{q})$. Finally, the sequence of reference benchmarks that we propose to use in this paper is $\mathbf{u}_t = \mathbf{q}^{\mathcal{H}}_i, \forall t \in \mathcal{T}^{\rm{oos}}_i$. We will empirically investigate the regret performance of OLNV in the case study presented in Section~\ref{sec:case_study}.

\section{Illustrative examples} \label{sec:illustrative_examples}

This section analyzes several illustrative examples to gain insight into the behavior of OLNV. The first case compares the two alternative implementations introduced in Section~\ref{subsec:ogd_implementation} and discusses their main properties. As a result of this analysis, we select the subgradient objective function as the default procedure to perform the update of $\mathbf{q}_t$ in OLNV. One of the key features of online learning algorithms is their tracking ability, given the chronological order in which the updates are performed. In the second illustrative example, we deal with alternating penalty scenarios, showing the salient properties of OLNV to adapt to a changing environment.

\subsection{Comparing the smooth and subgradient implementations}\label{subsec:toy_smooth}

This illustrative example aims to elucidate whether the smooth approximation presented in~\eqref{eq:nv_smooth_obj} provides any advantage over the direct subgradient implementation of OLNV. This will allow us to determine which implementation to be used for further numerical experiments. 

We consider a simplified setting with a single feature, a forecast of the wind power generation that we also use as the baseline for the FO method, and a single regressor $q_t \in \mathbb{R}$. No intercept is considered to ease the representation and analysis of $q_t$. We sample the feature from a uniform distribution $x_t \sim U[10, 90]$ (MW) and the true wind generation series is built adding a Gaussian noise, $E_t = x_t + \epsilon_t$ with $\epsilon_t \sim \mathcal{N}(0, 6)$ (MW). We generate a dataset of a 1-year duration (8760 samples, as if of hourly temporal resolution). Given that the penalties $\psi_t^{+}$ and $\psi_t^{-}$ are difficult to simulate, we compute them based on real day-ahead and regulation prices of the Danish DK1 bidding zone. We retrieve data corresponding to the year 2017 from the data portal of the Danish TSO, Energinet\footnote{See \url{https://www.energidataservice.dk/}}. Four implementations of Algorithm~\ref{alg:nv} are executed, three of them computing gradients of the smooth objective function through~\eqref{eq:nv_gradient} with $\alpha = 0.05$, $5$ and $20$ and the last one using subgradients of the original cost mapping as in~\eqref{eq:nv_subgradient}, to which we refer to as $\partial$. All instances are initialized with $q_1 = 1$, which means that the first offer produced by FO and OLNV are the same. In this section we do not use any convex transformation of the prices, i.e., $\mu = 1$, and we set $\eta = 0.005$. We run the OLNV algorithm throughout the dataset, performing updates of $q_t$ every hour.

In this section, we accompany the numerical results with some theoretical analysis. 
The function $NV_{t,\alpha}(\mathbf{q})$ approximates well the original function $NV_t(\mathbf{q})$ when $\abs{E_t - \mathbf{x}^\top_t \mathbf{q}} \rightarrow \infty$ as shown in Proposition~\ref{prop:nv_asymptotic} in Appendix~\ref{sec:appendixA}. Then, an interesting point of analysis related to the behavior of both functions in the neighborhood of $E_t - \mathbf{x}^\top_t \mathbf{q} = 0$, defined by $\varphi = \{\mathbf{q}: - \delta \le  E_t - \mathbf{x}^\top_t \mathbf{q} \le \delta \}$ with $\delta > 0$. Let $\mathbf{q}_1$ and $\mathbf{q}_2$ be two vectors with $E_t - \mathbf{x}^\top_t \mathbf{q}_1 \le 0$, $E_t - \mathbf{x}^\top_t \mathbf{q}_2 \ge 0$ and $\mathbf{q}_1, \mathbf{q}_2 \in \varphi$. The subgradient that OLNV computes for each vector changes substantially with $\partial NV_t(\mathbf{q}_1) = \psi^{-}_t \mathbf{x}_t$ and $NV_t(\mathbf{q}_2) = - \psi^{+}_t \mathbf{x}_t$, which may result in very different updates of the vector $\mathbf{q}$ for similar values of $\mathbf{x}_t$ or $\mathbf{q}_t$. Conversely, $NV_{t,\alpha}$ is everywhere differentiable, which ensures a smooth change of $\nabla NV_{t,\alpha}(\mathbf{q})$ for similar values of $\mathbf{q}-T$ and $\mathbf{x}_t$. 


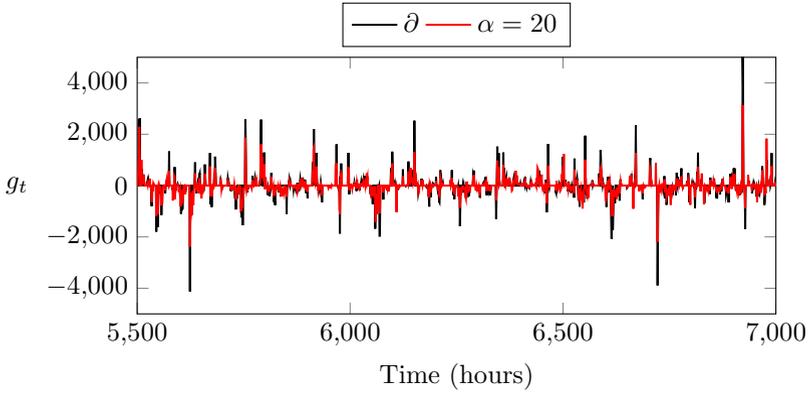
\begin{figure}
    \centering
    \begin{tikzpicture}
	\begin{axis}[	
    width=10cm, height=5cm,
    xmin = 5500, xmax = 7000, ymin=-5000, ymax=5000,
    xtick distance=500,
	legend style={at={(0.5,1.04)}, anchor=south, legend cell align=left, legend columns=2},
	clip marker paths=true,	
	xlabel = Time (hours),
	ylabel = $g_t$,
	ylabel style={rotate=-90},
	]
	 \addplot[line width=0.75pt, draw=black] table [x=index, y=subgra, col sep=comma] {data/toy1_gradient_evol.csv}; \addlegendentry{$\partial$}
	 \addplot[line width=0.75pt, draw=red] table [x=index, y=alpha20, col sep=comma] {data/toy1_gradient_evol.csv}; \addlegendentry{$\alpha=20$}
	\end{axis}	
    \end{tikzpicture} 
    \caption{Sample of $\partial NV_{t}$ and $\nabla NV_{t,20}$ computed in the dataset of the illustrative example.}  
    \label{fig:toy_gradient_evol}
\end{figure}





Figure~\ref{fig:toy_gradient_evol} shows a sample of $\partial NV_{t}$ and $ \nabla NV_{t,20}$ that corresponds to the subgradient and gradient of the smooth objective function with $\alpha = 20$. Only $NV_{t}$ and $NV_{t, 20}$ are represented, for the sake of clarity. Most of the spikes in the case of $NV_{t, 20}$ are comparatively lower due to the aforementioned smoothing effect in the neighborhood of $E_t - \mathbf{x}^\top_t \mathbf{q} = 0$. This is aligned with the decreasing value of the standard deviation of the (sub-)gradients $\sigma$ collated in Table~\ref{table:toy_smooth_q_evol} as $\alpha$ increases. 

\begin{table}[htp]
    \centering
    \caption{Average absolute value $\abs{\overline{g}}$ and standard deviation $\sigma$ of the (sub-) gradients and the metric $NV^{\rm{oos}}(\%)$ computed for three smooth ($\alpha$) and one subgradient ($\partial$) implementations of the OLNV. }
    \label{table:toy_smooth_q_evol}
    \begin{tabular}{ccccc}
        \hline
             & $\partial$ & $\alpha=0.05$ & $\alpha=5$ & $\alpha=20$    \\
        \hline
        $\abs{\overline{g}}$ & 121.7 & 122.0 & 125.6 & 133.5   \\
        $\sigma$             & 380.8 & 379.7 & 310.4 & 293.3   \\
        $NV^{\rm{oos}} (\%)$      & 5.3 & 5.2 & 0.8 & -14.5 \\
        \hline
    \end{tabular}
\end{table}

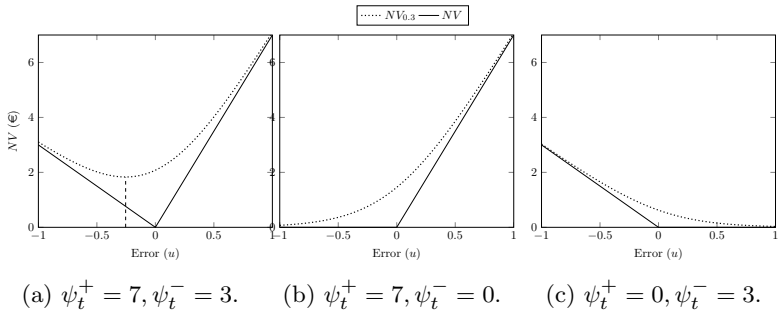
\begin{figure}
    \centering
    \captionsetup[subfig]{justification=centering}
    \begin{subfigure}[t]{0.28\textwidth}
        \centering
        \begin{tikzpicture}[scale=0.45]
            \pgfmathsetmacro{\aa}{0.3}
            \pgfmathsetmacro{\pp}{7}
            \pgfmathsetmacro{\pm}{3}
            \pgfmathsetmacro{\tt}{\pp / (\pp + \pm)}
            \pgfmathsetmacro{\xm}{\aa * ln ((1 - \tt) / \tt)}
            \pgfmathsetmacro{\ym}{\pp * \xm + \aa * (\pp + \pm) * ln( 1 + exp(-\xm / \aa))}
        	\begin{axis}[
            xmin=-1, xmax=1, ymin=0, ymax=\pp,
            xtick distance=0.5,
        	legend style={at={(1.6,1.05)},anchor=south,legend cell align=center,legend columns=3},	
        	samples=201,
        	clip marker paths=true,	
        	xlabel = Error ($u$),
        	ylabel = $NV$ (\euro),
        	]
        	\addplot[mark=none, dotted, domain=-1:1, line width=1.pt, draw=black] {\pp * x + \aa * (\pp + \pm) * ln( 1 + exp(-x / \aa))}; \addlegendentry{$NV_{0.3}$}
        	\addplot[mark=none, domain=-1:1, line width=0.75pt, draw=black] {max(- \pm * x, 0) + max(\pp * x, 0)};\addlegendentry{$NV$}
        	\addplot[mark=only, dashed] coordinates {(\xm, 0) (\xm, \ym)};
        	\end{axis}	
        \end{tikzpicture}
        \caption{$\psi^{+}_t = 7, \psi^{-}_t = 3$.}
        \label{fig:toy_nv_obj_func_pm}
    \end{subfigure}
    \begin{subfigure}[t]{0.28\textwidth}
        \centering
        \begin{tikzpicture}[scale=0.45]
            \pgfmathsetmacro{\aa}{0.3}
            \pgfmathsetmacro{\pp}{7}
            \pgfmathsetmacro{\pm}{0}
        	\begin{axis}[
            xmin=-1, xmax=1, ymin=0, ymax=\pp,
            xtick distance=0.5,
        	samples=201,
        	clip marker paths=true,	
        	xlabel = Error ($u$),
        	]
        	\addplot[mark=none, dotted, domain=-1:1, line width=1.pt, draw=black] {\pp * x + \aa * (\pp + \pm) * ln( 1 + exp(-x / \aa))}; 
        	\addplot[mark=none, domain=-1:1, line width=0.75pt, draw=black] {max(- \pm * x, 0) + max(\pp * x, 0)};
        	\end{axis}	
        \end{tikzpicture} 
        \caption{$\psi^{+}_t = 7, \psi^{-}_t = 0$.}
        \label{fig:toy_nv_obj_func_p}
    \end{subfigure}
    \begin{subfigure}[t]{0.28\textwidth}
        \centering
        \begin{tikzpicture}[scale=0.45]
            \pgfmathsetmacro{\aa}{0.3}
            \pgfmathsetmacro{\pp}{0}
            \pgfmathsetmacro{\pm}{3}
        	\begin{axis}[
            xmin=-1, xmax=1, ymin=0, ymax=7,
            xtick distance=0.5,
        	samples=201,
        	clip marker paths=true,	
        	xlabel = Error ($u$),
        	]
        	\addplot[mark=none, dotted, domain=-1:1, line width=1.pt, draw=black] {\pp * x + \aa * (\pp + \pm) * ln( 1 + exp(-x / \aa))};
        	\addplot[mark=none, domain=-1:1, line width=0.75pt, draw=black] {max(- \pm * x, 0) + max(\pp * x, 0)};
        	\end{axis}	
        \end{tikzpicture} 
        \caption{$\psi^{+}_t = 0, \psi^{-}_t = 3$.}
        \label{fig:toy_nv_obj_func_m}
    \end{subfigure}
    \caption{Different instances of the original $NV$ and smooth $NV_{0.3}$ objective function with $\alpha=0.3$ and $u = E_t - x_t q$.}
    \label{fig:toy_nv_obj_func}
\end{figure}

On the contrary, the mean absolute value of the (sub-)gradients, denoted as $\abs{\overline{g}}$, follows the opposite evolution. To understand the rationale behind this evolution, we provide Figure~\ref{fig:toy_nv_obj_func} showing three instances of the original and smooth losses. In all cases, we see that ${NV}_{t, \alpha}$ is an upper bound for ${NV}_{t}$ by a finite amount as expressed in Proposition~\ref{prop:nv_approximation} (with a proof available in Appendix~\ref{sec:appendixA}). However, Figure~\ref{fig:toy_nv_obj_func}(a) shows that the minimum of ${NV}_{t, \alpha}$ is not aligned with the minimum of the original pinball loss function. This is true whenever $\psi^{+}_t \neq \psi^{-}_t$ (i.e., asymmetric penalties in the market), a common situation in markets with a two-price imbalance settlement. Furthermore, when one penalty is equal to zero, the minimum is never attained.

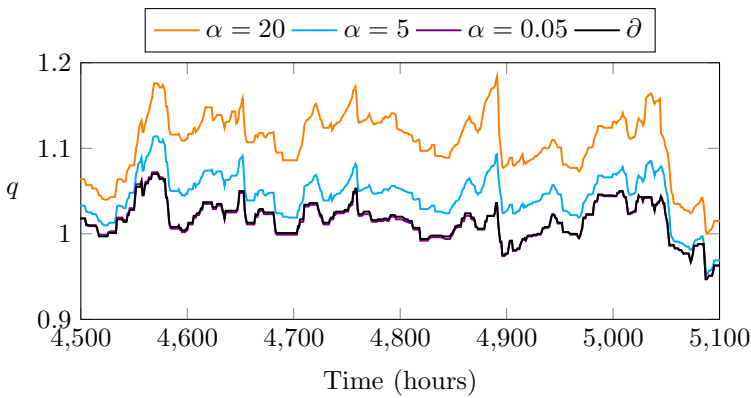
\begin{figure}
    \centering
    \begin{tikzpicture}
	\begin{axis}[	
    width=10cm, height=5cm,
    xmin = 4500, xmax = 5100, ymin=0.9, ymax=1.2,
	legend style={at={(0.5,1.04)}, anchor=south, legend cell align=left, legend columns=4},
	clip marker paths=true,	
	xlabel = Time (hours), ylabel = $q$,
	ylabel style={rotate=-90},
	]
	\addplot[line width=0.75pt, draw=orange] table [x=index, y=q_a20, col sep=comma] {data/toy1_q_evol.csv}; \addlegendentry{$\alpha=20$}
	\addplot[line width=0.75pt, draw=cyan] table [x=index, y=q_a5, col sep=comma] {data/toy1_q_evol.csv}; \addlegendentry{$\alpha=5$}
	\addplot[line width=0.75pt, draw=violet] table [x=index, y=q_a0.05, col sep=comma] {data/toy1_q_evol.csv}; \addlegendentry{$\alpha=0.05$}
	\addplot[line width=0.75pt, draw=black] table [x=index, y=q_sub, col sep=comma] {data/toy1_q_evol.csv}; \addlegendentry{$\partial$}
	\end{axis}	
    \end{tikzpicture} 
    \caption{Example of the evolution of the coefficient $q$ for different implementations of OLNV.} 
    \label{fig:toy_smooth_q_evol}
\end{figure}

Consequently, the gradient computed through \eqref{eq:nv_gradient} almost always introduces a deviation that is positive, compared to the true value returned by \eqref{eq:nv_subgradient}. The value of this error is given by the following expression:
%
\begin{align} \label{eq:nv_subgradient_diff}
   \nabla {NV}_{t, \alpha} - & \partial NV_t(\mathbf{q}) = \nonumber\\
   & \begin{cases}
         (\psi^{+}_t + \psi^{-}_t) (1 + e^{(E_t - \mathbf{x}^\top_t \mathbf{q}) / \alpha})^{-1}  \mathbf{x}_t, &  E_t - \mathbf{x}^\top_t \mathbf{q} > 0  \, , \\
        - (\psi^{+}_t + \psi^{-}_t) (1 + e^{-(E_t - \mathbf{x}^\top_t \mathbf{q}) / \alpha})^{-1}  \mathbf{x}_t, &  E_t - \mathbf{x}^\top_t \mathbf{q} < 0  \, , \\
        [-\frac{\psi^{+}_t + \psi^{-}_t}{2} \mathbf{x}_t, \frac{\psi^{+}_t + \psi^{-}_t}{2} \mathbf{x}_t], &  E_t - \mathbf{x}^\top_t \mathbf{q} = 0  \, . \\
    \end{cases}
\end{align}
The imperfect approximation of ${NV}_{t, \alpha}$ distorts the magnitude and even the sign of the gradients, causing a long-term drift of $q_t$ that increases with the smoothing parameter $\alpha$ as shown in Figure~\ref{fig:toy_smooth_q_evol}. 

Finally, the last row of Table~\ref{table:toy_smooth_q_evol} presents the $NV^{\rm{oos}}(\%)$ obtained by each implementation with respect to FO. One sees that $NV^{\rm{oos}}(\%)$ deteriorates when $\alpha$ increases. The smooth approach increasingly dampens the evolution of the decision vector for higher values of $\alpha$, but at the expense of a biased $q_t$ and with non-negligible economic losses. Therefore, the smooth approximation does not provide any substantial advantage over the subgradient implementation in this application, given that the producer is neutral to risk and volatility (only being concerned with expected profits), while there is no technical constraint that encourages a smooth evolution of $q$. As a consequence, we will only use subgradients to implement the OLNV method throughout the remainder of the manuscript.





\subsection{Dynamic behavior}\label{subsec:toy_dynamic}


In this illustrative example, we compare the tracking ability of OLNV and LP approaches in a nonstationary environment. Similar to the previous case, we assume that the producer has access to a unique feature and considers a model with a single regressor. Again, we sample the forecast from a uniform distribution $x_t \sim U(10, 90)$ (MW) and the true wind power generation series is obtained by adding a normal noise $E_t = x_t + \epsilon_t$ with $\epsilon_t \sim \mathcal{N}(0, 6)$ (MW). Instead of the real DK1 data, we consider two possible scenarios with penalties $\psi_t^+=1, \psi_t^-=3$ and $\psi_t^+=3, \psi_t^-=1$, alternating every two months. This process yields 8 months of data (5760 hours) using the last 4 months (2880 hours) as the test set. The start of the test set is aligned with the beginning of a two-month scenario with $\psi_t^+=1$ and $\psi_t^-=3$. The rolling window approach is implemented solving the optimization problem~\eqref{eq:nv_rolling_win_opt} with a set of historical samples $\mathcal{T}^{\rm{in}}$. Then, we use~\eqref{eq:nv_rolling_win_eval} to cast an offer based on the context $E^{\rm{F}}_t = \pi(x_{t}, q^{\text{LP}}_t)$. The coefficient $q^{\text{LP}}_t$ is refreshed every 24 hours by solving problem~\eqref{eq:nv_rolling_win_opt}, and based on a rolling window.  The reason for a 24-hour update is twofold: it is aligned with the original proposal in \cite{munoz2020feature} and we empirically checked that there was little economic gain to be obtained with more frequent updates. The computing time in the case of an hourly update, for example, took 24 times longer. As will be shown in the following, LP based on a rolling window approach only produces small changes over the training set, resulting in similar $q^{\text{LP}}_t$. We train four versions of the LP model with $\card{\mathcal{T}^{\rm{in}}} = 720$, $1440$, $2160$ and $2880$ (1, 2, 3, or 4 months), denoted as LP-1M to LP-4M, respectively. We use the first four months of the dataset to construct the initial training sets. Although the concept of training is not strictly the same for OLNV (since it always learns on the fly, as new samples become available), only the last month of the training set is used to update the value of $q_t$, originally initialized with $q_1 = 1$, to resemble a model that has been operating for some time. 

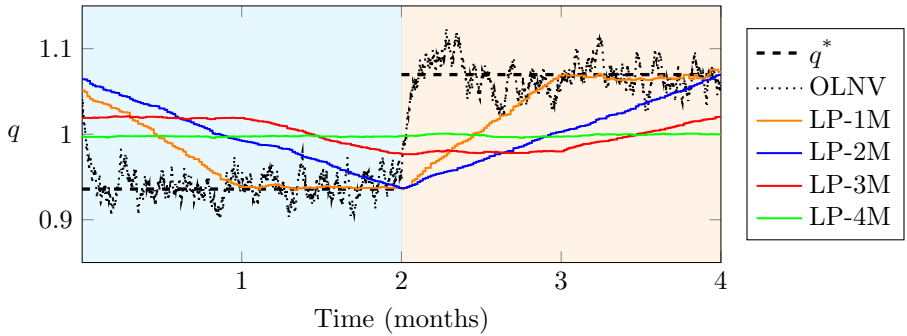
\begin{figure}
\centering
\begin{tikzpicture}
	\begin{axis}[	
    width=10cm, height=5cm, 
    xmin = 1, xmax = 2880, ymin=0.85, ymax=1.15,
    xtick={0, 720, 1440, 2160, 2880},
    xticklabels={0, 1, 2, 3, 4},
	legend style={at={(1.04,0.5)}, anchor=west, legend cell align=left, legend columns=1},
	clip marker paths=true,	
	xlabel = Time (months), ylabel = $q$,
	ylabel style={rotate=-90},
	]
	\draw [fill=cyan, opacity=0.1, draw=none] (axis cs:0,0.85) rectangle (axis cs:1440, 1.15);
	\draw [fill=orange, opacity=0.1, draw=none] (axis cs:1440,0.85) rectangle (axis cs:2880, 1.15);
	\addplot[mark=none, line width=1.25pt, dashed, draw=black, domain=0:1440, forget plot] {0.935899};
    \addplot[mark=none, line width=1.25pt, dashed, draw=black, domain=1440:2880] {1.06969}; \addlegendentry{$q^{\text{*}}$}
	\addplot[line width=0.75pt, draw=black, dotted] table [x=ts_index, y=q_online, col sep=comma] {data/toy2_q_evol.csv}; \addlegendentry{OLNV}
	\addplot[line width=0.75pt, draw=orange] table [x=ts_index, y=q_lp_m1, col sep=comma] {data/toy2_q_evol.csv}; \addlegendentry{LP-1M}
	\addplot[line width=0.75pt, draw=blue] table [x=ts_index, y=q_lp_m2, col sep=comma] {data/toy2_q_evol.csv}; \addlegendentry{LP-2M}
	\addplot[line width=0.75pt, draw=red] table [x=ts_index, y=q_lp_m3, col sep=comma] {data/toy2_q_evol.csv}; \addlegendentry{LP-3M}
 \addplot[line width=0.75pt, draw=green] table [x=ts_index, y=q_lp_m4, col sep=comma] {data/toy2_q_evol.csv}; \addlegendentry{LP-4M}
	\end{axis}	
    \end{tikzpicture}
\caption{Evolution of $q$ produced by five models over the test set. The blue and orange shaded periods correspond to the penalty scenarios $\psi_t^+=1, \psi_t^-=3$ and $\psi_t^+=3, \psi_t^-=1$, respectively. The entry $q^{*}$ corresponds to the best single vector for each penalty scenario.}   \label{fig:toy2_q_evolution}
\end{figure}

Figure~\ref{fig:toy2_q_evolution} depicts the evolution of the single regressor $q_t$ over the test set, together with the optimal $q^*$ for each penalty scenario. Over the first two months, the higher value of $\psi_t^-$ penalizes offers above the true production $E^{\rm{F}}_t > E_t$ and, consequently, the optimal strategy is to underestimate $E^{\rm{F}}_t$ with $q^* < 1$. Over the final months, we observe the opposite. 

As one may expect, the evolution of the decision vector of LP models is smoother than in the case of OLNV, given that the former approach considers many historical samples at once to perform the update. However, Figure~\ref{fig:toy2_q_evolution} also shows that the trajectory of $q_t$ produced by the rolling window models LP-1M to LP-4M is substantially lagged with respect to the change in the penalty scenario (emphasized by different background colors). This delay increases with the length of the training set, to the point that LP-4M completely overlooks it. Note that the length of the training set in LP-4M and the period of the penalty scenarios are identical. Therefore, the number of samples that penalizes under- or overproduction is equal and remains constant. As a result, LP-4M offers no incentive to overestimate or underestimate the forecast, yielding the same value as FO (neglecting slight deviations due to the finite sample and noise). 

Figure~\ref{fig:toy2_q_evolution} additionally shows that OLNV is substantially faster at tracking the optimal $q^*$. In contrast, the LP problem~\eqref{eq:nv_rolling_win_opt} determines the decision $q_t$ with the best performance on average in the training set, assuming that all the samples in the set are equally probable representations of future outcomes. Conversely, OLNV only uses the most recent information to perform a point-wise update that swiftly captures changes in the environment.

The tracking capability of both approaches has an impact on their economic performance. Table~\ref{table:toy_track_performance} summarizes the out-of-sample $NV^{\rm{oos}}(\%)$ obtained by each approach in the test set. In line with the previous analysis, LP-4M obtains the same performance as FO. The other three LP methods experience decreasing $NV^{\rm{oos}}(\%)$ as the length of the training set and the lag of $q_t$ increase. Finally, the adaptability of OLNV allows outperforming the LP approaches.

\begin{table}[htp]
    \centering
    \caption{Out-of-sample $NV^{\rm{oos}}$ (\%) obtained in the test set of the illustrative example.}
    \label{table:toy_track_performance}

    \begin{tabular}{cccccc}
    \hline
         &  OLNV & LP-1M  & LP-2M   & LP-3M   & LP-4M  \\
    \hline
    $NV^{\rm{oos}}$ (\%)  &  13 &  5 &  -5 & -6 &  0 \\
    \hline
    \end{tabular}
    
\end{table}

In this simplified example, we could have analyzed LP models with a shorter training set, probably resulting in reduced lag and better performances. However, in a realistic situation with a huge feature space and random penalties, months of data are typically required to capture the underlying relationships and generalize well in the out-of-sample set \citep{munoz2020feature}. Therefore, the length of the training set of the LP models has to be selected as a trade-off; enough data is required to learn a policy that generalizes well, but shorter sets capture dynamics better. On the contrary, the OLNV approach completely avoids this dichotomy, providing a fast and effective method that adapts to uncertain parameters generated by nonstationary environment.


\section{Case study} \label{sec:case_study}




Electricity markets are in the midst of a rapid development towards reducing the time between market transactions and the actual exchange of electricity. Examples of this transformation are given, i.e., by the reduction of the electricity lead time (Australian NEM or the Californian CAISO\footnote{See \url{https://aemo.com.au} and \url{http://www.caiso.com/}}) or by the development of new intraday markets (OMIE intraday markets or NordPool ELBAS\footnote{See \url{https://www.omie.es/} and \url{https://www.nordpoolgroup.com/}}). Inspired by this trend, we analyze a case study that considers an online forward market that takes place every hour followed by a balancing market with a two-price imbalance settlement. The gate closure of the forward market happens just before the start of the next period. We assume that the wind farm continuously participates in the market and her offer is always accepted.

In the following we first describe the data used in this case study. Then, several benchmark methods are proposed to compare against OLNV. Finally, in a last part, we analyze the numerical results obtained, based on regret, economic performance and computational costs.

\subsection{Data and experimental setup}

This case study is based on historical data compiled by the Danish TSO, Energinet.dk, since it includes market prices and several wind power forecasts that can be employed as input features. 
We collect the true and day-ahead forecast issued by Energinet for the on- and offshore wind power production of both DK1 and DK2 Danish bidding zones together with the day-ahead and regulation prices of DK1 for the period 01/07/2015 to 06/04/2021 (mm/dd/yyyy). The day-ahead spot and regulation prices are mapped into hourly penalties through equations~\eqref{eq:psi_pls} and \eqref{eq:psi_min} and some small negative values, obtained due to rounding errors, are filtered out.  


\begin{table}[htp]
    \centering
    \caption{Installed capacity in MW by bidding zone and technology.}
    \label{table:wind_capacity}

    \begin{tabular}{ccccc}
    \hline
         & \multicolumn{2}{c}{DK1} & \multicolumn{2}{c}{DK2}\\
    year & Onshore & Offshore & Onshore &  Offshore \\
    \hline
    2015 & 2966  & 843   & 608 & 428  \\
    2016 & 2966  & 843   & 608 & 428  \\
    2017 & 2966  & 843   & 608 & 428  \\
    2018 & 3664  & 1277  & 759 & 423  \\
    2019 & 3669  & 1277  & 757 & 423  \\
    2020 & 3645  & 1277  & 757 & 423  \\
    2021 & 3725  & 1277  & 756 & 423  \\
    \hline
    \end{tabular}
\end{table}

The raw wind power forecast series are also processed to suit our needs. Given that the installed capacity of the four wind categories shown in Table~\ref{table:wind_capacity} varies differently over the dataset, we independently normalize each series to lie between 0 and 100 MW, a figure that can easily represent the capacity of a large wind farm. According to the Danish TSO, the raw wind power forecasts are issued between 12 to 36 hours ahead, although the exact time is difficult to know because no timestamp is provided. To overcome this issue, we use a standard ordinary least square regression model to produce enhanced forecasts with an accuracy comparable to an hour-ahead forecast and, therefore, suitable for our case study. We feed each raw wind power forecast into an independent linear regression model together with the last three lags of the true historical wind realization of the pertaining series.
Finally, we use the first 6 months of our dataset to independently train each of the four predictive models, one per column of Table~\ref{table:wind_capacity}.

\begin{table}[htp]
    \centering
    \caption{Average RMSE (MWh) of the original forecast, the persistent (naive 1h lag) and improved 1h-ahead forecast computed on the out-of-sample period 07/01/2015 to 06/04/2021 with a normalized generation capacity of 100 MW.}
    \label{table:wind_mrse}

    \begin{tabular}{ccccc}
    \hline
    \multirow{2}{3em}{Model}   & \multicolumn{2}{c}{DK1} & \multicolumn{2}{c}{DK2} \\
           & Onshore &   Offshore & Onshore &  Offshore \\
    \hline
    original   & 6.19 & 9.55 & 6.77 & 10.68 \\
    persistent & 3.36 & 6.39 & 3.90 & 7.49  \\
    improved   & 2.72 & 5.70 & 3.34 & 6.66  \\
    \hline
    \end{tabular}
\end{table}

Table~\ref{table:wind_mrse} compares the root mean square error (RMSE) of the original and improved out-of-sample forecast against the naive benchmark provided by the first lag of each series (the wind power production of the previous hour), also known in the literature as persistence. Results show that the improved hour-ahead series significantly outperforms both original forecasts and persistence. As a byproduct, note that the wind power forecasts issued by the Danish TSO have quality metrics (e.g., RMSE) that are consistent with expectation, i.e., with offshore conditions being harder to predict than onshore conditions, while DK2 also having lower predictability since having small capacity and coverage area.

Once we have processed the wind power production series, we explain how we use them in our case study. The power generation of the wind farm offering in the market is simulated using the normalized onshore time-series of the Danish DK1 bidding zone, which is consistent with the bidding zone of the imbalance penalties utilized. The four hour-ahead forecasts of the wind power production of DK1 and DK2 are available to the producer as contextual information. Although additional wind power forecasts of neighboring bidding zones could have been used as features, we restrict ourselves to the ones produced by the Danish TSO to avoid potential inconsistencies regarding the issuing time that could cast doubt on the results obtained \citep{munoz2020feature}.

Given that our goal is to reduce the imbalance cost incurred by the wind farm, we also consider several price-related features to be used as contextual information. To this end, we include the first lag of the imbalance penalties $\psi^+_{t-1}$ and $\psi^-_{t-1}$ in the vector of contextual information. As commented in Section~\ref{sec:nv_rolling_win}, it is well known that the ratio between the penalties provides valuable information about the optimal decision of the newsvendor model and, therefore, we add the series $r_{t-1} = \psi^+_{t-1} / ( \psi^+_{t-1} + \psi^-_{t-1} + \upsilon)$ where $\upsilon = 10^{-5}$ is a constant that helps better condition the denominator. Finally, we add a column of ones that enable one of the regressors to become an intercept, completing our feature set. 

As a summary, let $E_t^{on1}, E_t^{of1}, E_t^{on2}, E_t^{of2}$ denote the hour-ahead wind power forecast of DK1 onshore, DK1 offshore, DK2 onshore and DK2 offshore, respectively. Then, at the moment of delivering the offer, the producer has available a feature vector $\mathbf{x}_t = [1, E_t^{on1}, E_t^{of1}, E_t^{on2}, E_t^{of2}, \psi^+_{t-1}, \psi^-_{t-1}, r_{t-1}]^\top$ to infer the optimal offer $E_t^{\rm{F}}$.



\subsection{Benchmark methods and implementation details}
\label{subsec:benchmark_implementation}


In this section, we describe several benchmark methods against which we compare the performance of OLNV. The first benchmark approach is the enhanced hourly forecast of DK1 itself, produced through the ordinary least square regression model described before. Although a prediction that minimizes the RMSE may seem naive, one can expect that the deviation cost incurred by the producer vanishes as the RMSE of the forecast approaches zero. Therefore, an hour-ahead forecast is expected to perform relatively well. We also use this hour-ahead forecast as the baseline to compute the metric $NV^{\rm{oos}}(\%)$ for the rest of the approaches in the way described in Section~\ref{subsec:performance_eval}.




The second benchmark is that of \cite{munoz2020feature}, based on two-step approach using two variants of \eqref{eq:nv_rolling_win_opt}. In the first step, the first model only considers wind-related features plus the intercept and set $\psi^{+}_t = \psi^{-}_t = 1, \enskip \forall t$. The series resulting from such model can be interpreted as an enhanced forecast of the wind energy production with a reduced mean absolute error. In a second step, this enhanced forecast is fed into~\eqref{eq:nv_rolling_win_opt}, considering this time the true historical penalties $\psi^{+}_t$ and $\psi^{-}_t$ to correct for market patterns but neglecting the capacity constraint \eqref{eq:nv_dis_feature2}. The training set is updated following a rolling window, adding new samples and eliminating the same amount of the oldest. We replicate this method, called LP2 (Linear Programming 2-steps), considering the four-hour-ahead enhanced wind power forecasts of DK1 and DK2 as the input of the first step, this is, $\mathbf{x}_t = [1, E_t^{on1}, E_t^{of1}, E_t^{on2}, E_t^{of2}]^{\top}$. In line with their findings, we choose a training set of $\card{\mathcal{T}^{\rm{in}}} = 4320$ (6 months) and a rolling window step of 24 hours. 


In addition, we analyze a rolling window model, called LP, that solves exactly \eqref{eq:nv_rolling_win_opt} and \eqref{eq:nv_rolling_win_eval} using the full vector of available contextual information. This method is the one from the illustrative example in Section~\ref{subsec:toy_dynamic}, but with different inputs. Given the similarities with the other rolling window approach LP2, we also choose a training set length of 6 months and a rolling window step of 24 hours. 

Finally, we discuss a benchmark that cannot be implemented in practice, inspired by the static regret metric defined in \eqref{eq:regret_nv_stdr}. We assume perfect information about the whole out-of-sample dataset and consider \eqref{eq:nv_rolling_win_opt} to compute the best linear model in hindsight, determined by the vector $\mathbf{q}^{\mathcal{H}}$. Once this optimal single vector is computed, the whole sequence of offers is determined through $E^{\rm{F}}_{t} = \pi(\mathbf{x}_{t}, \mathbf{q}^{\mathcal{H}})$. We name this benchmark FX (for FiXed). 

\textcolor{black}{Next, we discuss the implementation of OLNV in this case study. The OLNV algorithm does not need to solve an optimization problem but requires initializing two parameters. To choose $\mu$ and $\eta$, we perform an offline grid search on the chunk of data spanning 07/01/2015 to 12/31/2015. As candidate values for $\mu$ we consider $[0, 0.1, \hdots, 1]$ and for $\eta$ we analyze $[10^{-2}, 10^{-3}, 10^{-4}]$. The grid search is carried out executing $3 \times 11 = 33$ independent instances of the OLNV algorithm, initializing each time the OLNV regressor associated with the onshore DK1 forecast to $1$ and the rest of the values to $0.01$. The average $NV^{\rm{oos}}(\%)$ obtained by each instance is collated in Table 5. After analyzing the results, we select the combination of values $\mu = 0.7$ and $\eta = 0.001$ which achieve the highest $NV^{\rm{oos}}(\%)$. Even though in this case study a grid search was used for the sake of clarity, other more complex cross-validation techniques \citep{refaeilzadeh2009cross} can be used instead to select the values of $\mu$ and $\eta$, including repeating this process periodically to update the values of $\mu$ and $\eta$ after a change in the environment.}

In this case study, we assume a balanced penalty anchor $\overline{\psi}^+ = \overline{\psi}^- = 1$. Again, we initialize the OLNV regressor associated with the onshore DK1 forecast to $1$ and the rest of the values to $0.01$. In other words, we start the online offering with a strategy very close to FO, mainly relying on the forecast of the wind energy production. We use the next 6 months (01/01/2016 to 06/30/2016) to update (initialize) $\mathbf{q}_{\text{OL}}$ with the aim of having a fair comparison against LP and LP2.

The performance of all the methods presented in this section is evaluated using the dataset spanning from 07/01/2016 to 06/04/2021 (5 years with 43 200 samples). The optimization models LP, LP2, and FX are implemented with the Python package Pyomo \citep{bynum2021pyomo} and solved through the optimization solver CPLEX\footnote{IBM ILOG CPLEX Optimization Studio. See \url{https://www.ibm.com/analytics/cplex-optimizer}.}, whereas the implementation OLNV is developed by the authors based on standard Python packages and uploaded to an open repository\footnote{Experiment's code and data available at: \url{https://github.com/Miguel897/online-trading-wind-energy}}.

\begin{table}[htp]
    \setlength{\tabcolsep}{4pt}
    \centering
    \caption{Out-of-sample $NV^{\rm{oos}}$ (\%) for different combinations of parameters $\mu$ and $\eta_0$ over the span 07/01/2015 to 12/31/2015. Highlighted in black are shown the best result and parameters selected.}
    \label{table:cross_val}

    \begin{tabular}{c  ccccccccccc}
    \hline
    \multirow{2}{*}{$\eta$}  & \multicolumn{11}{c}{$\mu$}     \\
          & 0     & 0.1   & 0.2   & 0.3   & 0.4   & 0.5   & 0.6   & \textbf{0.7}  & 0.8  & 0.9   & 1     \\
    \hline
    $10^{-2}$          & -13,8 & 19,2 & 33,7 & 19,2 & 27,7 & 8,4  & 39,7 & 29,2 & 32,3 & 32,3 & 42,0 \\
    \textbf{$10^{-3}$} & 12,5  & 27,1 & 33,7 & 36,9 & 39,2 & 39,9 & 42,1 & \textbf{42,2} & 42,0 & 41,6 & 41,5 \\
    $10^{-4}$          & -5,2  & 1,3  & 4,4  & 6,0  & 7,0  & 7,7  & 8,2  & 8,6  & 8,9  & 9,4  & 9,4 \\
    \hline
    \end{tabular}
\end{table}


\subsection{Numerical results}

Next, we discuss the results obtained in this case study. We start examining the regret suffered by OLNV over the aforementioned out-of-sample dataset with a length of $D=43,200$ hours (60 months). Let $\mathcal{T}_j^{\rm{oos}} = \cup^{j}_{i=1} \mathcal{T}^{\rm{oos}}_i$ and recall $\mathbf{u}_t = \mathbf{q}^{\mathcal{H}}_i \enskip \forall t \in \mathcal{T}^{\rm{oos}}_i$. We assess the average dynamic regret $R_{T}^d/T$ for each sequence $\mathcal{T}_j^{\rm{oos}}, j=1,...,D/l$ with partition length $l=2160$, $4320$, $8640$ hours ($3$, $6$, $12$ months). As an additional case, we compute the evolution of the static regret for a sequence $\mathcal{T}_j^{\rm{oos}}, j=1,...,20$ with a step of $l=2160$ hours (3 months). In each step, we refresh the best single action in hindsight as $\mathbf{q}^{\mathcal{H}}_j = \argmin_{\mathbf{q} \in Q^{\mathcal{H}}_j} \sum_{t\in \mathcal{T}_j^{\rm{oos}}} NV_t(\mathbf{q})$ and $\mathbf{u}_t = \mathbf{q}^{\mathcal{H}}_j \enskip \forall t$.

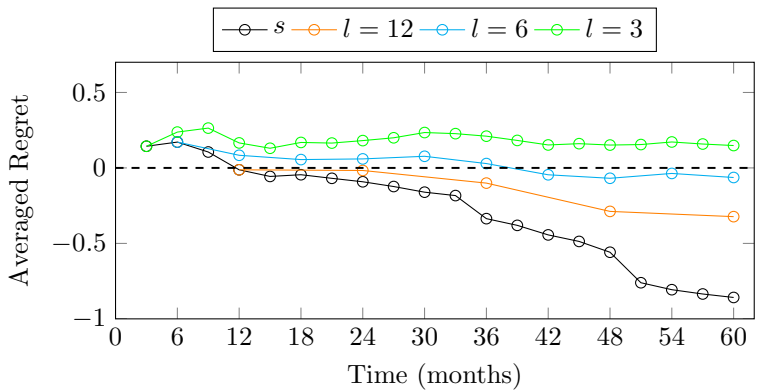
\begin{figure}
    \centering
    \begin{tikzpicture}
	\begin{axis}[	
    width=10cm, height=5cm, 
    xmin = 0, xmax = 62, ymin=-1, ymax=0.7,
    xtick distance=6,
    ytick={-1, -0.5, 0, 0.5},
	legend style={at={(0.5,1.04)}, anchor=south, legend cell align=left, legend columns=4},
	clip marker paths=true,	
	xlabel = Time (months), ylabel = Averaged Regret
	]
	 \addplot[draw=black, mark=o] table [x=months, y=3m_static, col sep=comma] {data/case_study_regret.csv}; \addlegendentry{$s$}
	 \addplot[draw=orange, mark=o] table [x=months, y=12m_dynamic, col sep=comma] {data/case_study_regret.csv};\addlegendentry{$l=12$}
	 \addplot[draw=cyan, mark=o] table [x=months, y=6m_dynamic, col sep=comma] {data/case_study_regret.csv}; \addlegendentry{$l=6$}
	 \addplot[draw=green, mark=o] table [x=months, y=3m_dynamic, col sep=comma] {data/case_study_regret.csv}; \addlegendentry{$l=3$}
	 \addplot[line width=0.75pt, mark=none, dashed, forget plot] coordinates {(0,0) (66,0)};
	\end{axis}	
\end{tikzpicture} \\
    \caption{Average dynamic regret $R^d_{T} / T$ for $l=3, 6, 12$, months and static regret $R^s_{T} / T$ updated every 3 months (denoted as $s$) of the OLNV method.}  
    \label{fig:case_study_regret}
\end{figure}

The four aforementioned regret series are depicted in Figure~\ref{fig:case_study_regret}. As expected, the average dynamic regret incurred by OLNV deteriorates quickly as $l$ decreases since lower values of $l$ translate in a more challenging benchmark closer to the the worst-case regret defined in \eqref{eq:regret_nv_worst}. Nevertheless, Figure~\ref{fig:case_study_regret} clearly shows that OLNV achieves a sublinear static regret, i.e., $\lim_{T \rightarrow \infty} \sup \mathcal{R}^s_T / T \le 0$. This is also the case for the dynamic regret with partitions of length $l \ge 6$ months, proving the ability of OLNV to track dynamic environments.

The economic gains obtained by each method are assessed through the $NV^{\rm{oos}}(\%)$. The  average values achieved over the evaluation dataset are collated in Table~\ref{table:cost_reduction}. First, note that all methods outperform the naive FO strategy of offering the DK1 forecast, obtaining positive values and demonstrating that this set of features contributes to reducing the deviation cost. 

The LP2 method is developed in a context where recent lags in the penalties are not available. Indeed, the lack of penalty-related features translates into a modest score, showing the evident benefits of disclosing recent information in electricity markets, i.e., reducing the lead time. Even though FX determines the optimal $\mathbf{q}^{\mathcal{H}}$ in hindsight  (i.e., under perfect information), its choice is limited to a single vector for the whole horizon. The fact that several approaches perform better than FX proves the dynamic behavior of the uncertain parameters and the need for updating the decision vector. Therefore, it does not come as a surprise that LP improves the first two approaches as it relies on the full vector of features and periodically updates $\mathbf{q}^{\text{LP}}_t$. However, the superior adaptability of OLNV allows it to obtain the best score, achieving an additional 7.6\% compared to LP and a total 38.6\% deviation cost reduction compared to FO. The latter figure translates into an extra 25,930.22 \euro/year on average for a wind farm with a capacity of 100 MW. 

Finally, the last row of Table~\ref{table:cost_reduction} summarizes the computational time corresponding to four approaches. The FX method requires little time as it only solves a single optimization problem for the whole horizon. This contrasts with the significant amount of time required by the constant re-optimization of LP and LP2. It is noteworthy that even though OLNV produces 24 times more updates of the vector $\mathbf{q}_t$, the time invested is several orders of magnitude lower. In conclusion, OLNV is up to the challenge of the electricity markets transformation achieving significant cost reduction together with exceptional computational performance.

\begin{table}[htp]
    \centering
    \caption{Out-of-sample $NV^{\rm{oos}}$ (\%) and execution time (s) over the span 07/01/2016 to 06/04/2021.}
    \label{table:cost_reduction}

    \begin{tabular}{cccccc}
    \hline
         & LP2  & FX   & LP  & OLNV \\
    \hline
    $NV^{\rm{oos}}(\%)$ &   3.8 & 24.6 & 31.0    & 38.6   \\
    Time (s)       & 23366 &   53 & 16077   & 179    \\
    \hline
    \end{tabular}
\end{table}


\section{Conclusions}\label{sec:conclusions}



 
This paper develops a new algorithm, named OLNV, combining a variant of the online gradient descent with recent advances that extends the newsvendor model to consider contextual information directly. The component-wise update of the learning rate enables the use of features with different scales seamlessly. In nonstationary environments, conventional stochastic approaches may consider misleading old samples in their training sets. On the contrary, our algorithm tracks the most recent information of the gradients, adapting the learning rate to follow the dynamics of the uncertain parameters and potentially obtaining higher profits. 
The closed-form expressions derived to compute the projection into the feasible region and a gradient of the objective function yield a efficient algorithm that can be used in computationally expensive problems. We envision the use of OLNV in future electricity markets that evolves toward continuous offering with reduced lead time. In particular, we apply this algorithm to the wind farm problem offering in an hourly forward market with a dual-price settlement for imbalances.


Several numerical experiments are carried out to assess the properties of the proposed OLNV algorithm. In the first illustrative example, we compare the behavior of two alternative implementations, namely, a subgradient approach and a smooth approximation of the original newsvendor function. The numerical and theoretical analysis provided in this example indicates that computing subgradient on the original objective function proves more profitable since it avoids update errors that may be introduced by the smooth approximation. Consequently, we determined that the subgradient implementation was the most suitable to this application and used it throughout the rest of the numerical experiments. Nevertheless, the smooth approximation could be utilized in other applications where other technical concerns advice a smooth update. 
 
The second example shows the adaptability of the OLNV algorithm to nonstationary environments, clearly outperforming other stochastic approaches that optimize (using mathematical programming techniques) over a training set of past information. This superior performance is justified by the point-wise update that only uses the most recent information. Our case study, built upon real data of the Danish TSO Energinet, shows that OLNV achieves a 38.6\% cost reduction against using a point forecast as offer and 7.6\% compared to a state-of-the-art method. These significant improvements contribute to accelerating the integration of renewable energy technologies. Furthermore, we empirically analyze several dynamic definitions of regret, showing the desired sublinear convergence against most benchmarks. 

\textcolor{black}{Although this research focused on wind energy producers, OLNV is readily applicable to managing a portfolio of variable renewable energies with zero marginal cost, including wind, solar and other technologies. Similar algorithms can be developed when the producer's portfolio includes other assets such as loads, thermal power plants, or energy storage facilities, replacing the aggregated source of uncertainty, i.e., the variable net energy production, by a linear decision rule. In this case, the projection step on the feasible region would likely involve solving a quadratic optimization program that can still be efficiently solved with modern solvers, when the feasible region is convex. Another attractive front is extending the OLNV algorithm to address inter-temporal constraints, observing a similar note with regard to the feasible region as in the previous case.  This may require first generalizing the newsvendor framework to offering in electricity markets though.}

Future work also includes delving into the theoretical guarantees that this algorithm offers in terms of regret. On a different front, a wealth of other algorithms within the field of online learning can be applied to this problem, potentially bringing additional benefits such as faster convergence rates or improved performance. Similarly, variable selection techniques could help determine the subset of the available feature streams that provide the most economic value, whereas nonlinear mapping, i.e., kernels or generalized additive models (GAMs), can extend the regression capabilities of the method. Another exciting line of research concerns the risk analysis of the producer, where other metrics can be used instead of the expected value to create risk-averse strategies.

\section*{Acknowledgments}

M. Á. Mu\~{n}oz is funded by the Spanish Ministry of Science, Innovation and Universities through the State Training Subprogram 2018 of the State Program for the Promotion of Talent and its Employability in R\&D\&I, within the framework of the State Plan for Scientific and Technical Research and Innovation 2017-2020 and by the European Social Fund.

P. Pinson and J. Kazempour are partly supported through the Smart4RES project (European Union’s Horizon 2020, No. 864337). The sole responsibility of this publication lies with the authors. The European Union is not responsible for any use that may be made of the information contained therein.



\begin{appendices}
\section{Smooth function properties} \label{sec:appendixA}
This appendix provides a lemma and several propositions related to the smooth approximation $NV_{t, \alpha}$ defined in~\eqref{eq:nv_smooth_obj}. Some of the proofs in this appendix are based on the proofs provided in \cite{zheng2011gradient}. 
In this appendix we assume that $\psi^{+}_t, \psi^{-}_t \ge 0$ and $\psi^{+}_t + \psi^{-}_t > 0 \enskip \forall t$. Next, we define an auxiliary function $S_{t, \alpha}(u)$, $S_{t, \alpha}: \mathbb{R} \rightarrow \mathbb{R}$ with $\alpha > 0$ as follows
%
\begin{align}
    S_{t, \alpha}(u) = \psi^{+}_t u + \alpha (\psi^{+}_t + \psi^{-}_t) \log (1 + e^{ - u / \alpha})  \, ,  \label{eq:aux_eq}
\end{align}
where $u \in \mathbb{R}$. We use this function in the proofs covered within this appendix. First, we prove the convexity of $S_{t, \alpha}$. 

\begin{lemma}\label{lemma:aux_eq_convexity}
For any given $\alpha > 0$, the function $S_{t, \alpha}$, defined in \eqref{eq:aux_eq}, is a convex function.
\end{lemma}

\noindent \emph{Proof} From the definition of $S_{t, \alpha}$ in~\eqref{eq:aux_eq}, we calculate that
\begin{align}
    \frac{d^2S_{t, \alpha}(u)}{du^2} = \frac{\psi^{+}_t + \psi^{-}_t}{\alpha} \frac{e^{-\frac{u}{\alpha}}}{(1 + e^{-\frac{u}{\alpha}})^2} > 0  \, ,
\end{align}
for any $u \in \mathbb{R}$ since $\psi^{+}_t + \psi^{-}_t >0$ and $\alpha > 0$.

We use this intermediate result to prove the convexity of $NV_{t, \alpha}$ in the following Proposition.

\begin{prop} \label{prop:smooth_nv_convexity}
For any given $\alpha > 0$, the function $NV_{t, \alpha}$, defined in \eqref{eq:nv_smooth_obj}, is a convex function of $\mathbf{q}$.
\end{prop}

\noindent \emph{Proof} Let $u = E_t - \mathbf{x}^\top_t \mathbf{q}$ in \eqref{eq:aux_eq}. Thus,
\begin{align}
    NV_{t, \alpha}(\mathbf{q}) = S_{t, \alpha}(E_t - \mathbf{x}^\top_t \mathbf{q})  \, . \label{eq:nv_equal_aux}
\end{align}
For $0 \le \omega \le 1$ and any $\mathbf{q}_1$ and $\mathbf{q}_2$, we have
\begin{align}
    NV_{t, \alpha}(\omega \mathbf{q}_1 + (1 - \omega) \mathbf{q}_2) &= S_{t, \alpha}(E_t - \mathbf{x}^\top_t (\omega \mathbf{q}_1 + (1 - \omega) \mathbf{q}_2)) \nonumber \\
    &= S_{t, \alpha}(E_t - \omega \mathbf{x}^\top_t \mathbf{q}_1 - (1 - \omega) \mathbf{x}^\top_t \mathbf{q}_2) \nonumber \\
    &= S_{t, \alpha}(\omega (E_t - \mathbf{x}^\top_t \mathbf{q}_1) - (1 - \omega)(E_t - \mathbf{x}^\top_t \mathbf{q}_2)) \\
    &\le \omega S_{t, \alpha}(E_t - \mathbf{x}^\top_t \mathbf{q}_1) + (1 - \omega) S_{t, \alpha}(E_t - \mathbf{x}^\top_t \mathbf{q}_2)  \, , \label{eq:nv_smooth_conv_prop_eq1} 
\end{align}
where the inequality in~\eqref{eq:nv_smooth_conv_prop_eq1} follows from the convexity of $S_{t, \alpha}$, proved in Lemma~\ref{lemma:aux_eq_convexity}. Then, using~\eqref{eq:nv_equal_aux}, the above inequality renders
\begin{align}
    NV_{t, \alpha}(\omega \mathbf{q}_1 + (1 - \omega) \mathbf{q}_2) \le \omega NV_{t, \alpha}(\mathbf{q}_1) + (1 + \omega) NV_{t, \alpha}(\mathbf{q}_2)  \, , \label{eq:nv_smooth_conv_prop_eq2}
\end{align}
showing that $NV_{t, \alpha}$ is a convex function on $\mathbf{q}$.

Next, we show that $NV_{t, \alpha}$ asymptotically approaches $NV_t$ for $\alpha \rightarrow 0$. We also show that the function $NV_{t, \alpha}$ upper bounds $NV_t$ for all $q \in \mathbb{R}^p$.

\begin{prop}\label{prop:nv_approximation}
Let $NV_t$ and $NV_{t, \alpha}$ be the functions defined in~\eqref{eq:nv_ob_online} and \eqref{eq:nv_smooth_obj}, in that order, with $\alpha > 0$. Then, we have
\begin{align}\label{eq:prop_2a}
    0 < NV_{t, \alpha}(\mathbf{q}) - NV_{t}(\mathbf{q}) \le  \alpha (\psi^{+}_t + \psi^{-}_t) \log 2  \, ,
\end{align}
for any $\mathbf{q} \in \mathbb{R}^p$. Thus,
\begin{align}\label{eq:prop_2b}
    \lim_{\alpha \rightarrow 0^+} NV_{t, \alpha}(\mathbf{q}) = NV_{t}(\mathbf{q})  \, .
\end{align}
\end{prop}

\noindent \emph{Proof} When $E_t - \mathbf{x}^\top_t \mathbf{q} \ge 0$, we have that 
\begin{align} \label{eq:nv_approx_ge_zero}
    NV_{t, \alpha}(\mathbf{q}) - NV_{t}(\mathbf{q}) = \alpha  (\psi^{+}_t + \psi^{-}_t) \log (1 + e^{-(E_t - \mathbf{x}^\top_t \mathbf{q}) / \alpha})  \, ,
\end{align}
hence,
\begin{align}
    0 < NV_{t, \alpha}(\mathbf{q}) - NV_{t}(\mathbf{q}) \le \alpha  (\psi^{+}_t + \psi^{-}_t) \log 2  \, ,
\end{align}
for $E_t - \mathbf{x}^\top_t \mathbf{q} \ge 0$. When $E_t - \mathbf{x}^\top_t \mathbf{q} < 0$,
\begin{align}
    NV_{t, \alpha}(\mathbf{q}) - NV_{t}(\mathbf{q}) &= (\psi^{+}_t + \psi^{-}_t) (E_t - \mathbf{x}^\top_t \mathbf{q}) \phantom{\alpha (\psi^{+}_t + \psi^{-}_t)  \log (1 + e^{-(E_t - \mathbf{x}^\top_t \mathbf{q}) / \alpha})} \nonumber \\
    & \hspace{10mm} + \alpha (\psi^{+}_t + \psi^{-}_t)  \log (1 + e^{-(E_t - \mathbf{x}^\top_t \mathbf{q}) / \alpha}) \\
    &= \alpha (\psi^{+}_t + \psi^{-}_t)  \log (1 + e^{(E_t - \mathbf{x}^\top_t \mathbf{q}) / \alpha})  \, . \label{eq:nv_approx_le_zero}
\end{align}
While 
\begin{align}
    0 < \alpha & (\psi^{+}_t + \psi^{-}_t) \log (1 + e^{(E_t - \mathbf{x}^\top_t \mathbf{q}) / \alpha}) \nonumber \\ 
     &< \alpha (\psi^{+}_t + \psi^{-}_t) \log (1 + e^{0 / \alpha}) = \alpha (\psi^{+}_t + \psi^{-}_t) \log 2  \, , 
\end{align}
since $E_t - \mathbf{x}^\top_t \mathbf{q} < 0$. This shows that $NV_{t, \alpha}(\mathbf{q}) - NV_{t}(\mathbf{q})$ also falls in the range $(0, \alpha (\psi^{+}_t + \psi^{-}_t) \log 2)$ for $E_t - \mathbf{x}^\top_t \mathbf{q} < 0$. Thus,~\eqref{eq:prop_2a} is proved. Eq.~\eqref{eq:prop_2b} follows directly by letting $\alpha \rightarrow 0^+$ in~\eqref{eq:prop_2a}.

Finally, we show that for high values of $\abs{E_t - \mathbf{x}^\top_t \mathbf{q}}$ the function $NV_{t, \alpha}$ asymptotically approximate  $NV_{t}$.

\begin{prop}\label{prop:nv_asymptotic}
Let $NV_t$ and $NV_{t, \alpha}$ be the functions defined in~\eqref{eq:nv_ob_online} and \eqref{eq:nv_smooth_obj}, in that order, with $\alpha > 0$. Then, when $\abs{E_t - \mathbf{x}^\top_t \mathbf{q}} \rightarrow \infty$ we have that  $NV_{t, \alpha} - NV_t \rightarrow 0$.
\end{prop}

\noindent \emph{Proof} For the sake of a clearer exposition we define $\mu(\mathbf{q}) = E_t - \mathbf{x}^\top_t \mathbf{q}$, where $\mu: \mathbb{R}^p \rightarrow \mathbb{R}$. When $\mu(\mathbf{q}) \rightarrow +\infty$, we have that
\begin{align}  \label{eq:nv_asymptotic_pos}
    \lim_{\mu(\mathbf{q}) \rightarrow +\infty} NV_{t, \alpha}(\mathbf{q}) - NV_{t}(\mathbf{q}) = \lim_{\mu(\mathbf{q}) \rightarrow +\infty} \alpha  (\psi^{+}_t + \psi^{-}_t) \log (1 + e^{-(\mu(\mathbf{q})) / \alpha}) = 0
\end{align}
When $\mu(\mathbf{q}) \rightarrow -\infty$, and using~\eqref{eq:nv_approx_le_zero}, we have that
\begin{align} \label{eq:nv_asymptotic_neg}
    \lim_{\mu(\mathbf{q}) \rightarrow -\infty} NV_{t, \alpha}(\mathbf{q}) - NV_{t}(\mathbf{q}) = \lim_{\mu(\mathbf{q}) \rightarrow -\infty} \alpha  (\psi^{+}_t + \psi^{-}_t) \log (1 + e^{(\mu(\mathbf{q})) / \alpha}) = 0
\end{align}
Combining both cases, this proposition is proved.





\end{appendices}

\bibliography{bibliography2}

\end{document}